\documentclass[final,authoryear,5p,times, twocolumn]{elsarticle}

\usepackage{multirow}
\usepackage{booktabs}
\usepackage[table,xcdraw,dvipsnames]{xcolor}

\usepackage{lineno}
\usepackage{graphicx}
\usepackage{color}
\usepackage{amssymb}
\usepackage{amsmath}
\usepackage{bigints}
\usepackage[bookmarksdepth=3]{hyperref}
\usepackage{lineno}
\usepackage{mathtools}
\usepackage{blkarray}
\usepackage[capposition=bottom]{floatrow}
\usepackage{varwidth}
\usepackage{cuted}

\usepackage{pifont}
\graphicspath{{img/}}
\usepackage{booktabs}

\usepackage{pgf,tikz}
\usepackage{mathrsfs}
\usetikzlibrary{arrows}
\hypersetup{
    colorlinks=true,
    citecolor=red,
    linkcolor=blue,
    filecolor=blue,      
    urlcolor=red,}
    
\journal{Cities}

\begin{document}

\begin{frontmatter}

\title{Can Machine Learning Uncover Insights into Vehicle Travel Demand from Our Built Environment?}

\cortext[cor1]{Corresponding author.}

\author[add1]{Zixun Huang\corref{cor1}}
\ead{zixun@berkeley.edu}

\author[add4]{Hao Zheng}

\address[add1]{University of California at Berkeley, CA, USA}
\address[add4]{City University of Hong Kong, Hong Kong, China}


\begin{abstract}
In this paper, we propose a machine learning-based approach to address the lack of ability for designers to optimize urban land use planning from the perspective of vehicle travel demand. Research shows that our computational model can help designers quickly obtain feedback on the vehicle travel demand, which includes its total amount and temporal distribution based on the urban function distribution designed by the designers. It also assists in design optimization and evaluation of the urban function distribution from the perspective of vehicle travel. We obtain the city function distribution information and vehicle hours traveled (VHT) information by collecting the city point-of-interest (POI) data and online vehicle data. The artificial neural networks (ANNs) with the best performance in prediction are selected. By using data sets collected in different regions for mutual prediction and remapping the predictions onto a map for visualization, we evaluate the extent to which the computational model sees use across regions in an attempt to reduce the workload of future urban researchers. Finally, we demonstrate the application of the computational model to help designers obtain feedback on vehicle travel demand in the built environment and combine it with genetic algorithms to optimize the current state of the urban environment to provide recommendations to designers.

\end{abstract}



\end{frontmatter}

\section{Introduction}

\subsection{Background}
A significant number of empirical studies demonstrate the interdependence between land use and local vehicle travel demand through household surveys \citep{cervero1991land, cervero1996mixed, cervero2003walking}. For example, researchers have shown that mixed-use development (MXD) promote the alternative role of non-motorized travel choices such as walking and cycling \citep{cervero1997travel, maria1997travel, park2018impacts}, and high density facilitates the use and development of public transit mode \citep{cervero1994transit, dunphy1996transportation}. And recent researchers have revealed that both the abundance and spatial configuration of urban land use are correlated with traffic congestion\citep{wang2021urban}.\par
Management of traffic resources using an urban design perspective proves to be feasible and effective \citep{cervero1997travel, mcnally1997assessment}. These urban studies receive significant attention to quantify and measure travel behavior, built environment, and the correlation between the two \citep{song2013comparing, ewing2001travel, clifton2008quantitative, ewing2010travel}. To encourage efficient utilization of traffic resources and improve the relation between traffic supply and demand, there exists tools for policymakers and urban designers to consider involving traffic analysis and prediction from the perspective of land use \citep{wachs1989regulating}.\par

\subsection{Problem Statement}
However, deficiencies and omissions in past studies hamper more detailed conclusions that help urban designers determine the relationship between vehicle travel demand and urban functions and then to carry out a better urban design.\par
The prevailing measurement methods of urban environment used in these studies omit specific information regarding urban function ratio, thus failing to provide urban designers with more detailed references. The entropy index, for example, illustrates this shortcoming. Different proportions of urban functions may produce the same entropy value, and the highest one hinders the designers in crafting the most appropriate LMU strategy \citep{song2013comparing}. Occasionally, the variances in research subjects, such as neighborhoods, are consolidated into a single categorical variable with a concomitant loss of information \citep{ewing2001travel}, including but not limited to function distribution information. However, some researchers have studied the influence of specific urban functions such as offices, residences, retail and transit on vehicle travel behaviors and demonstrated the impact of specific urban functions on transportation travel \citep{cervero2006reduces,  cambridge1994effects, choi2021utility}.\par
Limited information and dimensionality of survey data left a dearth of detail on the impact of urban function distribution on local vehicle travel demand temporal distribution. Peak congestion serves as a major challenge for urban traffic networks with limited bearing capacity. Average distribution of the traffic in time can help share the traffic pressure \citep{loudon1988predicting, downs2005still, shallal1980predicting}. Cervero suggested in his early research that trips tend to be more evenly distributed throughout day and week in a more mixed-function community \citep{cervero1996mixed, cervero1989land}. While most literature in this field mainly focus on the impact of land use on the total amount of traffic, such as vehicle miles traveled (VMT) and vehicle hours traveled (VHT), few studies focused on and made conclusions the relationship between urban land use and temporal distribution of vehicle travel demand.\par
Empirical evidence struggles to be repeated in new study areas, cultures, and eras. Traditional travel behavior modeling relies on labor-intensive, high-cost, and time-consuming data collection methods such as American Household Surveys and local travel surveys that make large scale, in-time analysis extremely difficult to achieve \citep{li2020understanding}. Past empirical evidence on the transportation impact description of built environments rendered itself inconsistent \citep{cervero1996mixed, cervero2006reduces} and unsuitable for comparison due to lacking reliable standard error estimates from individual studies \citep{ewing2010travel}. For instance, Cervero utilized building height as a proxy for employment density \citep{cervero1991land} while others use gross population density \citep{dunphy1996transportation} to measure a similar problem. Thus, designers often spend high labor costs to verify the practical effects of the empirical conclusions in a different region.\par
In summary, researches based on the traditional survey method have provided abundant conclusions of built environment factor, including but not limited to local density, diversity, and single-function distribution, and its impact on travel behavior. Designers find inspiration based on these findings to  construct a better planning decision. However, these research methods lack further detail regarding the spatial patterns of urban functions as well as the temporal patterns of vehicle travel demand. They also fail to illustrate the relationship between them and produce viable conclusions in other different areas.\par

\subsection{Data-driven and Machine Learning Based Urban Study}
With the development of artificial intelligence (AI) and urban information management, urban designers acquire opportunities to enhance traditional workflow and analyze in detail the traditional urban problem (i.e., how urban function distribution affects vehicle travel demand) from integrated quantitative and qualitative perspectives. Rapid advancement in information and communications technology has facilitated a generation of data capturing human movements and daily activities \citep{li2020understanding, chauhan2016addressing}. An increasing number of urban researchers have adopted ubiquitous urban data collected in real-time \citep{kang2020urban, shen2016urban, gervasoni2016framework, yuan2012discovering} to quantify and describe urban patterns and processes. Additionally, machine learning (ML) related algorithms have been used to create models that perform predictive functions in data-driven urban study \citep{nosratabadi2019state, souza2019data}. The consensus states that ubiquitous urban data combined with ML technology bears tremendous potential in the field of urban studies and smart city solutions \citep{nosratabadi2019state, hancke2013role, ibrahim2020understanding, hadjimichael2016machine, youssef2020machine}.\par
When applied, ML algorithms are capable of automating urban tasks such as land use and land cover (LULC) mapping and zoning \citep{demir2018deepglobe, kandrika2008land}; predicting future urban evolution \citep{grekousis2013modeling}; mapping residential density patterns jointly with multi-temporal Landsat data \citep{mccauley2004mapping}; discovering function regions with human mobility and points of interest (POI) data \citep{yuan2012discovering}; identifying land-use type based on temporal traffic patterns with taxi data \citep{liu2012urban}; automatically
evaluating the urban visual environment in a large scale \citep{liu2017machine}; estimating health outcomes at a neighborhood scale \citep{feng2021predicting}.\par
Related research on the application of ML in travel behavior includes applying decision tree (DT) induction \citep{loh2011classification} to predict transport mode \citep{wets2000identifying} and route choice \citep{yamamoto2002drivers, arentze2000using}. ML can also be used for exploring the relationship between ride-sourcing services and vehicle ownership \citep{sabouri2020exploring}. Deep learning approaches to predict the short-term demand of bike-sharing FFBS with weather and air quality data \citep{bao2019short} and to develop a prediction system for real-time parking in smart cities \citep{vlahogianni2016real}.\par

\subsection{Objective}
This research aims to develop a method capable of creating a computational model that maps detailed built environment information into vehicle travel demand information including temporal distribution and total amount. Such a model, therefore, aids urban designers in receiving feedback quickly and decision-makers in optimizing the allocation of resources to achieve a balance between traffic and other urban purposes like residence, commerce, industry, and infrastructure.\par
This research combines travel behavior data and urban function distribution to generate a machine learning model that offers vehicle travel demand prediction. We developed an ML model that learns experiences from urban data to help designers automate detailed traffic demand judgments in a certain area of a given city. Regarding the problem statement, the prediction model of this study is based on three setups.\par

\begin{enumerate}
  \item An input feature about local built environment defined as a ratio between urban functions in the region of study instead of a single categorical variable or a measure of land use mix;\par
  \item An output feature about travel behavior defined as the average daily quantity and temporal distribution of vehicle travel demand in study region instead of VMT or VHT;\par
  \item Commonly used open-source urban data utilized as training data sets for the prediction model to examine the cost efficiency of applying this research method to different cultural areas.\par
\end{enumerate}

This research aspires to use a computer, which inputs detailed urban function distributional information to the program, to predict the average daily quantity and temporal distribution of vehicle travel demand of a given local area. The temporal distribution image of vehicle travel demand becomes an output regarded as a reference to adjusting urban function distribution within the area. Then, the ML model is reused to verify the adjusted urban function distribution and determine whether the revision achieves a purpose that makes more rational use of traffic resources. Furthermore, urban function composition information can be iteratively optimized by combining the prediction model with a genetic algorithm.\par



\section{Methodology}
\begin{figure*}[!htb]
    \includegraphics[width=0.7\textwidth]{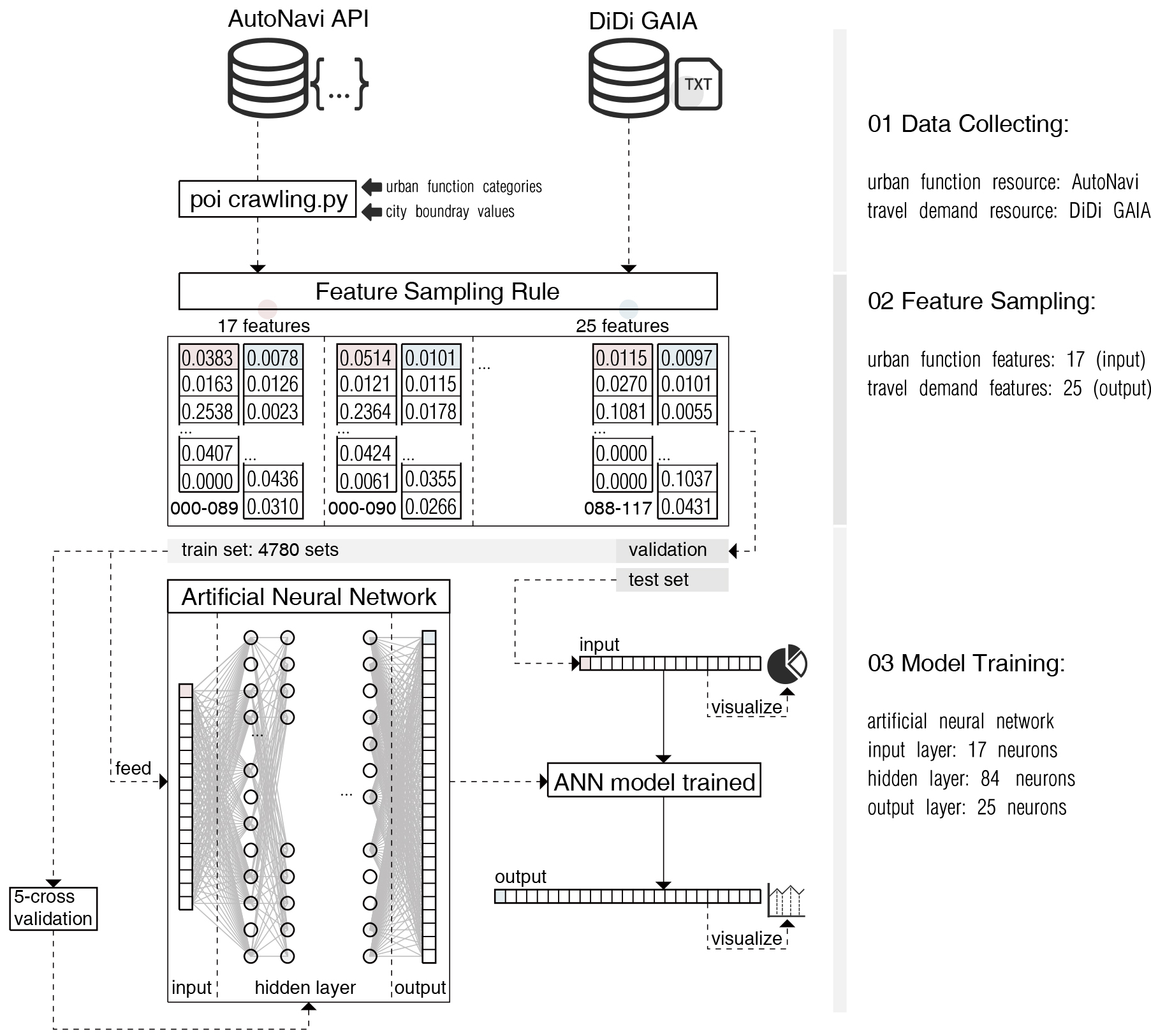}
    \caption{Data pre-processing and neural network training method.}
    \label{F25}
\end{figure*}

This section explains the method, primarily regarding data preparing, sampling, and model training. Figure \ref{F25} shows a general introduction to the production of a more detailed forecast of vehicle travel demand. First, data about the built environment and the travel demand should be collected and quantified through a web crawling and data cleaning process. After that, the data-set is redivided into a series of data samples based on geographic location. Subsequently, forecasting models such as Artificial Neural Networks map the relationship between the urban environment and vehicle travel demand based on the processed samples. The prediction models should serve as a guide for designers and decision-makers to interpret environmental information and meet the crowd travel demands.\par

\subsection{Data Preparing and Feature Sample Rule}
To describe the urban local function environment in detail, points of interest (POI) are selected as elements that form our living environment. Urban function density and mix in local areas of different scales, such as in function zones \citep{yao2017sensing}; in neighborhoods \citep{yue2017measurements}; in living areas \citep{liu2020spatial}; in 1km buffers \citep{zhao2018uncertain}; and at the census tract level \citep{maharana2018use}, are measured as the quantitative relationship between POI. In this study, we describe our urban function environment in the following way: the total number of POI types defined as $k$; the amount of POI type $j$ in a local area $Z$ denoted as $x_Z^j$; the density of all urban function types within the district $Z$ measured by proxy as $X_Z=\sum_{j=1}^{k} x_Z^j$; and the percentage of the urban function $j$ measured as $p_Z^j=\frac{x_Z^j}{X_Z}$. It follows that $\sum_{j=1}^{k} p_Z^j=1$.\par
To obtain an urban function spatial distribution containing latitudinal and longitudinal information, we choose AutoNavi as the data source, which is a Chinese web mapping, navigation and location-based services provider. Based on POI code table provided by AutoNavi, urban functions are divided into 16 categories (Table \ref{table:1}).\par

\begin{figure*}[!htb]
    \includegraphics[width=0.80\textwidth]{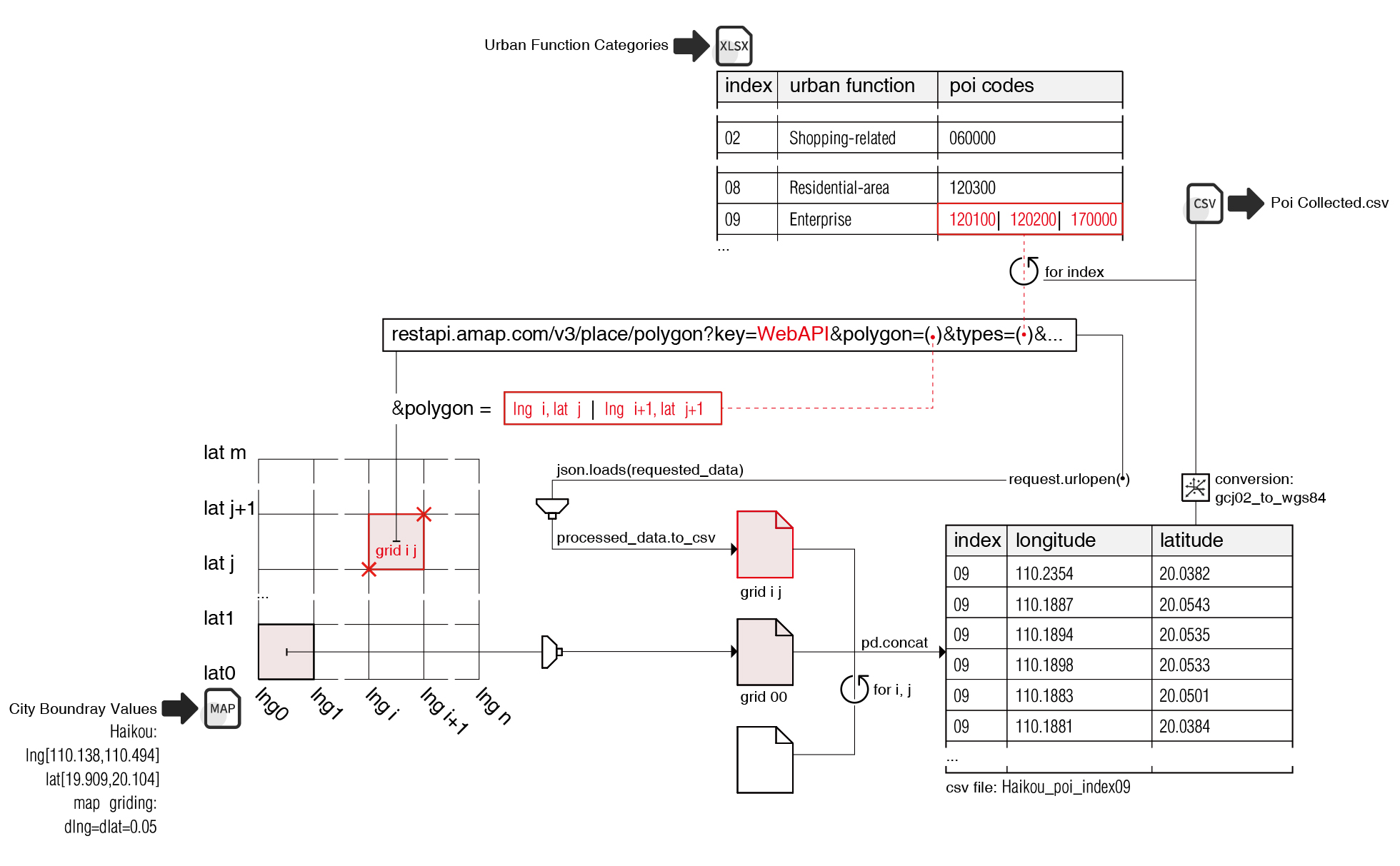}
    \caption{The process of built environment information crawling}
    \label{F21}
\end{figure*}

According to the API user guide, city code and POI code are initially selected as main crawling input parameters. However, we discovered that POI in each category are limited to 900 points with city code as a parameter input, which causes a sparse urban function distribution and leads to unreasonable equaling of different POI types. The urban function environment cannot be described accurately in this situation. To solve this problem, the entire target area is divided into smaller equal parts for finer crawling to acquire information on the built environment that more accurately reflects reality. The POI information in each part is then captured and categorized by POI types (Figure \ref{F21}).\par

After obtaining the detailed urban function distribution data, we sample the whole urban area uniformly for input features for machine learning later. The feature sampling rule follow as such (Figure \ref{F22}):

\begin{figure*}[!htb]
    \includegraphics[width=0.80\textwidth]{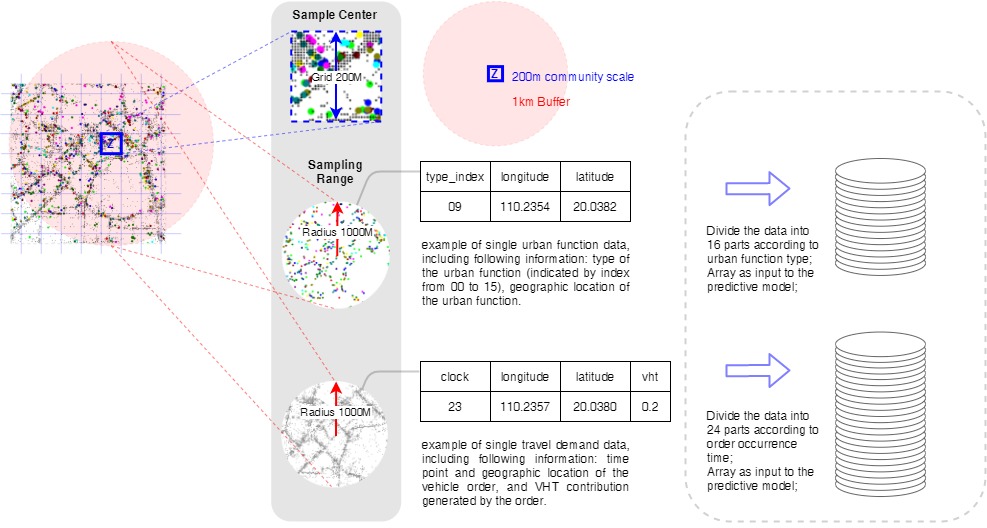}
    \caption{The flow chart of the feature sampling rule}
    \label{F22}
\end{figure*}

\begin{enumerate}
    \item A buffer area with a radius of 1 km, a standard unit of measurement in urban researches \citep{zhao2018uncertain, frank2004obesity, frank2005linking, sallis2016physical} regarding the relationship between land use and travel behavior, defined as a basic sampling unit; 
    \item The whole area of interest sampled every 200 m (i.e., general land block scale); 
    \item A sampling unit defined as region $Z$, the proportion of urban functions $p_Z^j$ and density $X_Z$, and measured as urban environment features of this unit. In addition, we normalize the density feature of each sample. The maximum density proxy value of all cells is denoted as $D$, and the normalized density value of each sample is redefined as $X_Z^*=\frac{X_Z}{D}$. The total number of samples is defined as $n$. For sample $Z$ ($Z$=1-$n$), the final 17 built environment features obtained by this sampling method are $X_Z^*$ and $p_Z^j$ ($j$=1-16), which describe the whole density of urban functions and detailed proportion of 16 urban functions in the study sample $Z$.\par
\end{enumerate}

Pertaining to vehicle travel demand data, this research adopts open-source data. Taxi as a door-to-door, all-weather way of travel becomes the subject of study. Vehicle travel demand in a certain place can be represented by taxi orders or vehicle hours traveled (VHT) generated by those orders. The regression model developed by Liu \citep{liu2020spatial} shows strong links between demand for taxis and urban form characteristics. For a large body of emergent data related to travel behavior \citep{liu2012urban, davis2016multi, tong2017simpler, zhang2017revealing}, taxi data proves to be an excellent option because of its ease of acquisition and large data size. Valuable actual and general data are generated along due to a mass of real behaviors. Moreover, vehicle hours traveled (VHT), which has been proven significant in environment research, is ultimately selected as the fundamental travel variable as a result of convincing, prior empirical studies by Ewing and Cervero about processing travel behavior data. \citep{ewing2001travel}.\par
Traffic order data are distributed evenly to each day and then divided into 24 intervals based on the time of order. Subsequently, these data are sampled in the same way mentioned above. In addition, information on hours traveled is extracted from the orders to describe exact travel demand based on Cervero's previous research. Therefore, each piece of order is bound with an additional weight coefficient representing VHT generated by order data. For a geographic unit $Z$ sampled, the average daily order is counted as $c_Z=\sum VHT$. The orders between $i$ o’clock and $i+1$ o’clock are formulated as $c_Z^i$ ($i$=0-23). The order distribution ratio in each interval is marked as $q_Z^i=\frac{c_Z^i}{c_Z}$. Therefore, it follows that $\sum_{i=0}^{23} q_Z^i=1$; For sample $Z$, the final 25 travel demand features obtained by this sampling method are normalized $c_Z^*$ and $q_Z^i$ ($i$=0-23), which describe the whole density of travel demands and detailed VHT distribution throughout a day in study sample $Z$.\par

\begin{figure}[!h]
    \includegraphics[width=1\textwidth]{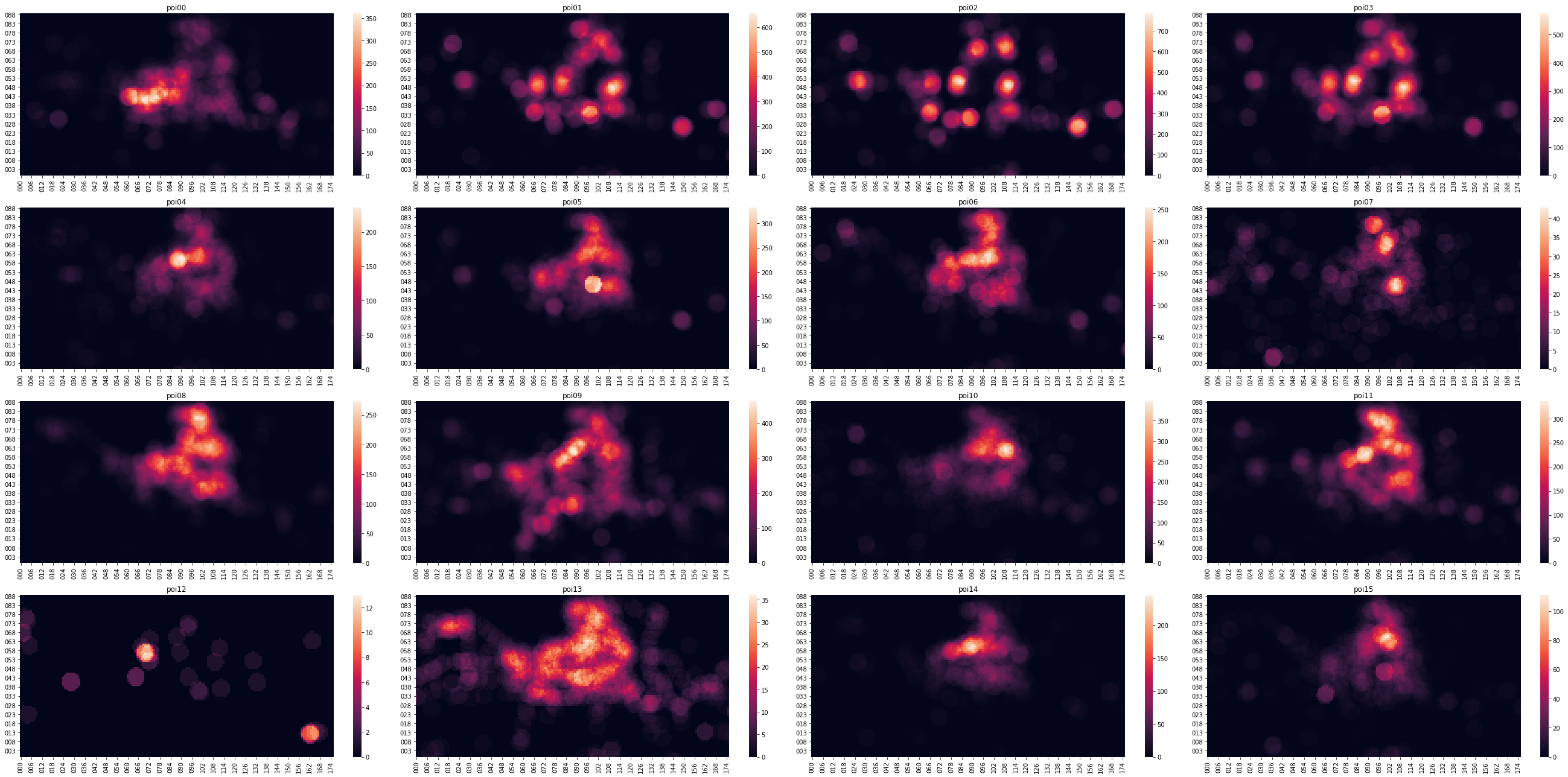}
    \caption{Heat distribution maps of 16 kinds of urban functions (sorted by Table \ref{table:1}). For each map, the brighter grid region contains the more function points.}
    \label{F23}
\end{figure}

\begin{figure}[!h]
    \includegraphics[width=1\textwidth]{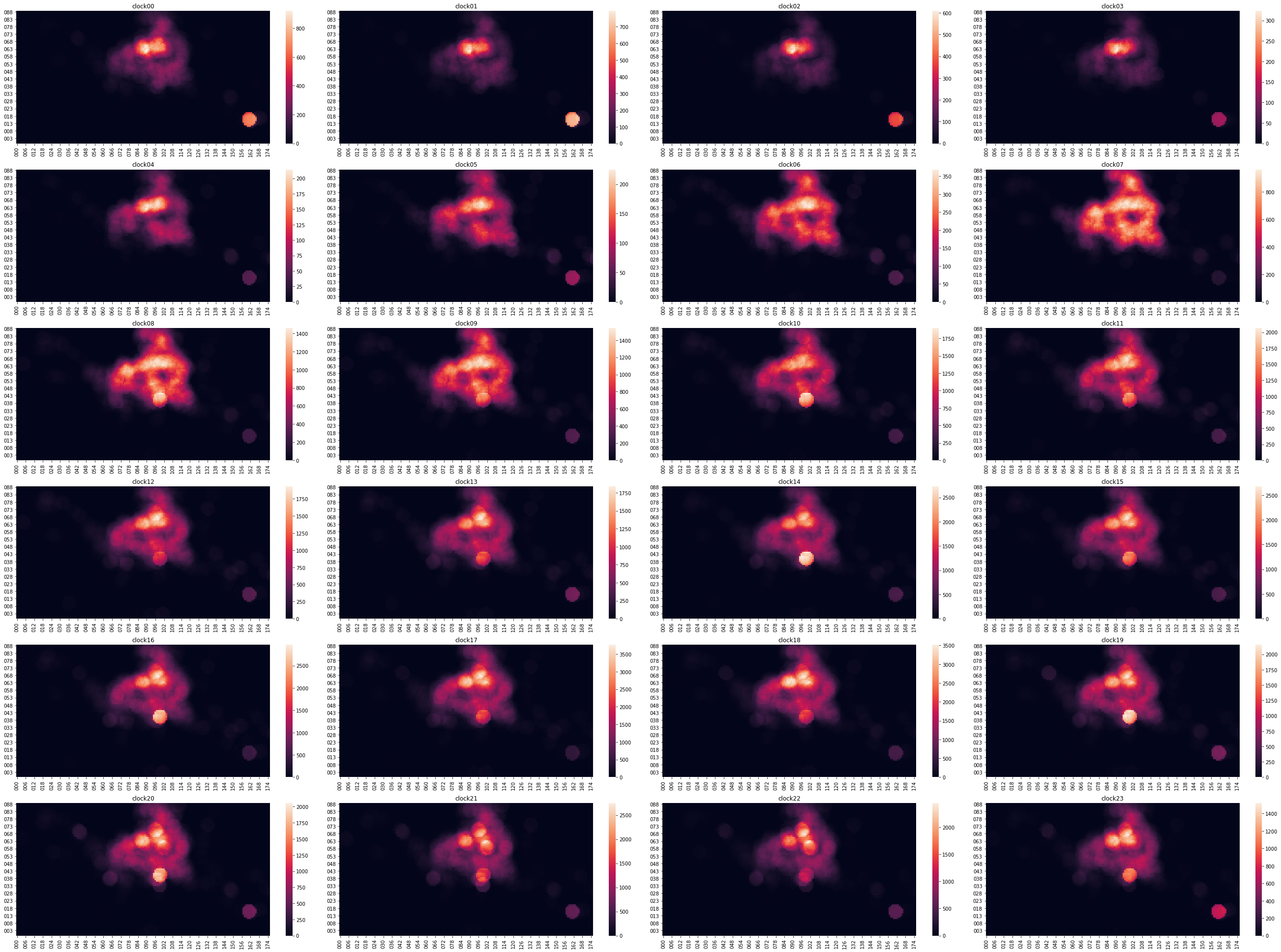}
    \caption{Heat distribution maps of 24 segments of VHT (from 0 o'clock to 23 o'clock). For each map, the brighter grid area represents the higher total VHT value compared with other grids.}
    \label{F24}
\end{figure}

\subsection{Study Area}
A port city in China named Haikou recently approved international tourism island construction. Consequently, a rapid increase of vehicle ownership and urban road traffic gradually became obstacles restricting the development of its social economy. The high sand content due to Haikou's special geology poses the risks of surface subsidence and seawater back-up. Thus, it is difficult to establish a subway system as public transportation to relieve the pressure of traffic.\par
A total of 65680 POI data of Haikou is obtained through a Python program called API provided by AutoNavi. A month’s amount of taxi order data in Haikou totaling 1,809,517 is provided by Didi Chuxing GAIA Initiative, an open-source data set. According to feature sampling rules, 15664 sample units were collected from the location of $(110.138,19.909)$ to $(110.494,20.104)$ in the GCJ02 coordinate system. Python is used to grid the geographical scope of Haikou and to calculate 17 built environment features and 25 travel demand features for each grid in this process. \par
Only 5991 valid units are reserved for neural network training after data cleaning operations to ensure that all samples contain valid built environment information and effective travel requirements. Samples that fail to contain any POI information or exceed one taxi order per hour on average in its radiation buffer are excluded.\par
In addition, (Figure \ref{F23}) and (Figure \ref{F24}) illustrate the urban function and travel data demand respectively collected in Haikou. Although not the focus of this research, they help us better understand the characteristics of the site in order to evaluate the prediction results of the computational model. Through these two sets of maps, we macroscopically see the distribution of regional centers of different functions and how active vehicle travel demand flows over time.\par

\subsection{Neural Network Training and Accuracy Definition}

This research endeavors to formulate computational models that predict vehicle travel demand that also include information about traffic temporal distribution with respect to the density and composition of urban functions. The prediction models should therefore help designers adjust and optimize their urban planning. \par

ANNs, Random Forest (RF), and Support Vector Machine (SVM) are chosen as the algorithms to obtain a computational model applicable to this study. ANN-based models, in particular, utilize Back-Propagation (BP), which fine-tune the weights of neurons to find practically accurate solutions for formulated problems and phenomena only understandable through experimental data and field observations \citep{basheer2000artificial}. They fulfill the problem of connections between our built environment and vehicle travel demand.\par

Understanding the complexity in both forecasting total demand and acquiring detailed information, we constructed different structures of neural networks referred to as network T and D. The built environment information processed by feature sampling rules can be initially set as 17 input neurons with values ranging from 0 to 1, indicating a proportion of 16 urban functions and a normalized density compared with the densest area. As for the output structures, the number of neurons is determined by the number of parameters needed to predict different network structures. For network T, total vehicle travel demand is set as one output neuron with a value from 0 to 1, indicating the normalized total demand compared with the most active area. For network D, detailed temporal distribution is set as 24 output neurons with a value from 0 to 1, indicating a 24-hr VHT distribution ratio. In addition, sigmoid function and mean squared error (MSE) are selected as activation and loss functions when training our neural networks.\par
The number of hidden layers and neurons therein should also be defined. Based on the literature \citep{karsoliya2012approximating} and actual tests, suitable results are obtained when the number of neurons of each hidden layer is 36 for network T and 82 for D. Five-fold cross-validation tests are employed to find neural network structures catering to the complexity of our study problem. Accuracy functions are customized to ascertain the loss function and help us better evaluate the accuracy of the outputs.\par
Equation \ref{Accuracy of density} is used to evaluate the prediction accuracy of the vehicle travel demand density (the total VHT amount all over the day). In this formula, we take the absolute value of the difference between the predicted value of the normalized local total amount of VHT and its ground-truth value as the error value.\par

\begin{equation}
Accuracy=1-\frac{\lvert groundtruth_{allday}-prediction_{allday} \rvert}{groundtruth_{allday}}
\label{Accuracy of density}
\end{equation}

Equation \ref{Accuracy of distribution} evaluates the prediction accuracy of detailed VHT temporal distribution. With it, we take the sum of the absolute value of the difference between the predicted values of VHT per hour and their ground truth value as error.\par

\begin{equation}
Accuracy=1-\sum_{clock=0}^{23}\lvert groundtruth_{clock}-prediction_{clock} \rvert
\label{Accuracy of distribution}
\end{equation}


\begin{table*}[!htb]
\begin{tabular}{|c c c|}
\hline
ANN Structure & Median accuracy for detailed temporal distribution (\%) & Median accuracy for local demand density (\%) \\
\hline
2- Layer ANN                                                     & 95.86                                                   & 91.6                                          \\
3- Layer ANN                                                     & 96.52                                                   & 93.86                                         \\
4- Layer ANN                                                     & 97.07                                                   & 94.1                                          \\
5- Layer ANN                                                     & 97.57                                                   & 94.81                                         \\
6- Layer ANN                                                     & 97.48                                                   & 95.52                                         \\
7- Layer ANN                                                    & 97.78                                                   & \textbf{95.68}                                         \\
8- Layer ANN                                                     & \textbf{97.84}                                                   & 95.22                                         \\
9- Layer ANN                                                     & 97.57                                                   & 95.1                                          \\
\hline
Others & - & - \\
\hline
Random Forest                                                    & 91.78                                                   & 91.88                                          \\
Linear SVR                                                    & 92.96                                                  & 91.61                                          \\
\hline
\end{tabular}
\caption{Five-fold cross-validation of neural networks with different numbers of layers.}
\label{table:2}
\end{table*}

We tested our neural networks on parts of the Haikou region using the processed environment information as the inputs and the processed demand data as the outputs. A total of 898 feature samples are divided into 5 parts, four for training and one for validation. Table \ref{table:2} presents the median accuracy of neural networks with different layer sizes in the fivefold cross-validation test. We also compared our artificial neural networks with other types of machine learning models such as the RF and Linear SVM. ANN noticeably achieves higher accuracy at least for predictions in the local area with randomly distributed samples. Meanwhile, the artificial neural network T with 7 layer-size sports the highest accuracy in the vehicle travel demand prediction of local total demand while the neural network D with 8 layers is most accurate in the demand prediction of detailed temporal distribution. As a result, the neural networks with 7 layer-size and 8 layer-size are chosen as the preliminary settings. In addition, we used a batch size of 100 in the preliminary settings and the customized error rate (1- accuracy) with increasing epochs recorded during the experiments to help us find an appropriate number of iterations in prediction model training (Figure \ref{F26}).\par

\subsection{Post Process and Visualization}
Using the trained model, the predictions of vehicle travel demand information are remapped into values before normalization. They respectively represent the total VHT values throughout one day and the proportions of VHT produced in 24 one-hour periods on geographic units. In addition, the prediction values are then visualized and compared with the ground truth values.


\section{Result and Application}
\subsection{Accuracy of Training Set and Test Set}
\begin{figure}[!h]
    \includegraphics[width=1\textwidth]{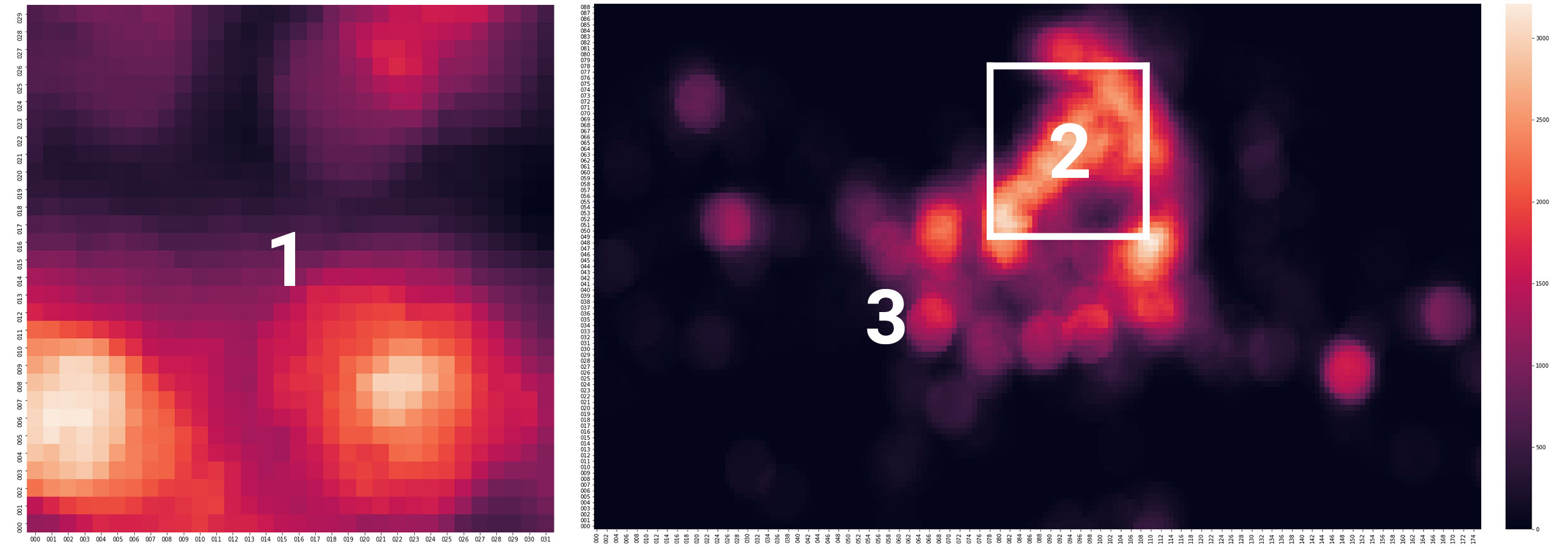}
    \caption{Urban function density map of the three different areas used in the accuracy test (2,3) and the multi-area hybrid test (1,2,3). (1: Chengdu Urban Area, 2: Haikou Urban Area, 3: the Entire Haikou)}
    \label{F30}
\end{figure}

Based on feature sampling rules and the training method settings mentioned above, two forecast models at various regional scales are built and trained with different data sets. Up to 5991 valid samples, inclusive of the built environment and vehicle travel demand information, represent the whole area of Haikou. 898 samples signifying the Haikou urban area (Figure\ref{F30}) are used as data sets to test the accuracy of the two neural networks. In the case of random geographic distribution, 80 percent of samples are used in the training process, and the remaining samples are used in accuracy tests.\par

\begin{figure}[!h]
    \includegraphics[width=0.97\textwidth]{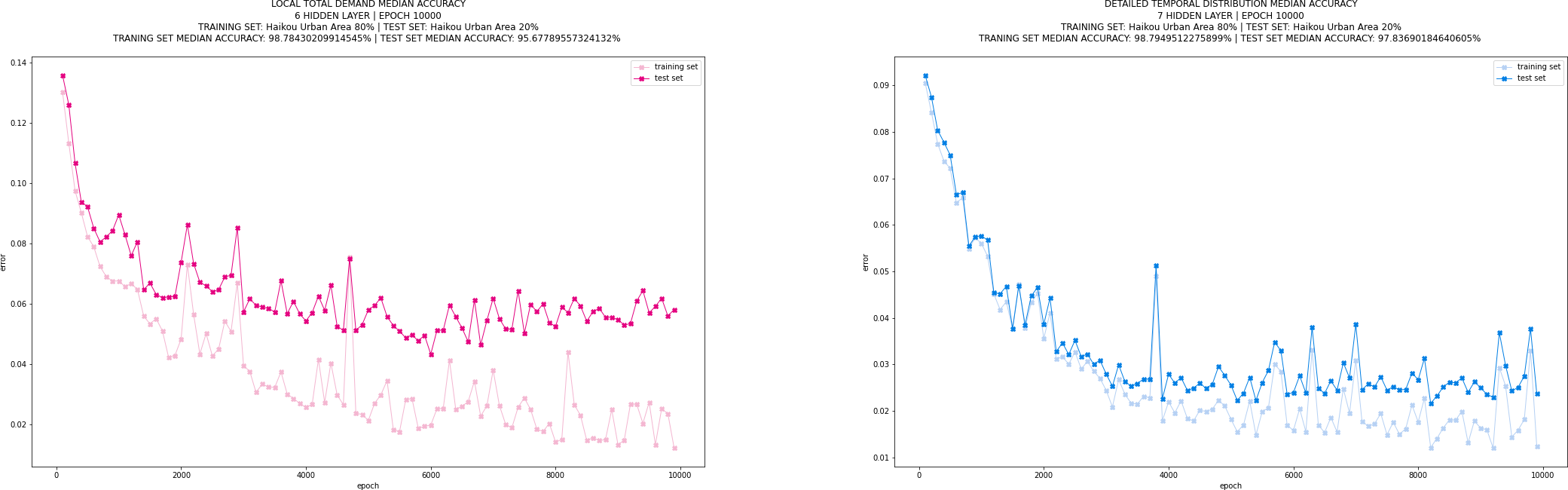}
    \caption{This figure presents the error rate variation of the prediction model in the urban area of Haikou with the number of iterations increasing. The red lines represent total VHT amount prediction error rates and the blue lines represent VHT temporal distribution prediction error rates; the bright lines represent the error rates variation in the test set, and the faded lines represent the error rates variation in the training set.}
    \label{F26}
\end{figure}

\begin{figure}[!h]
    \includegraphics[width=1\textwidth]{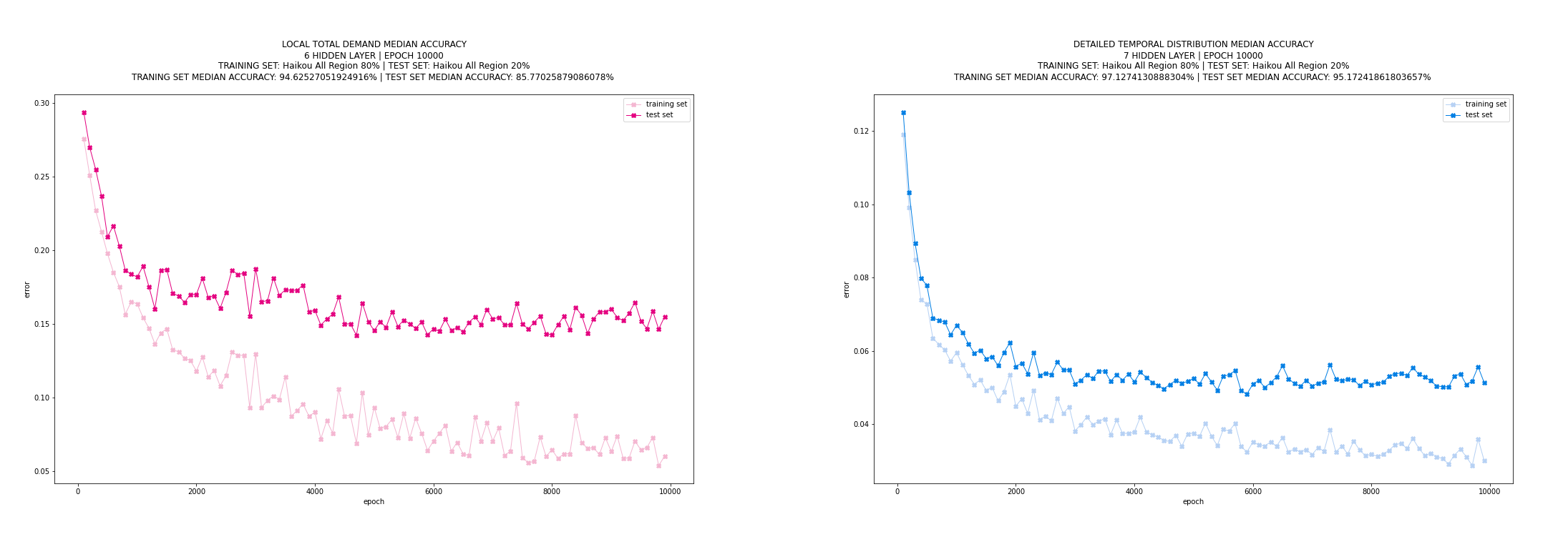}
    \caption{The same as Figure \ref{F26}. This figure presents the error rate variation of the prediction model in the whole area of Haikou with the number of iterations increasing.}
    \label{F27}
\end{figure}

\begin{figure}[!h]
    \includegraphics[width=1\textwidth]{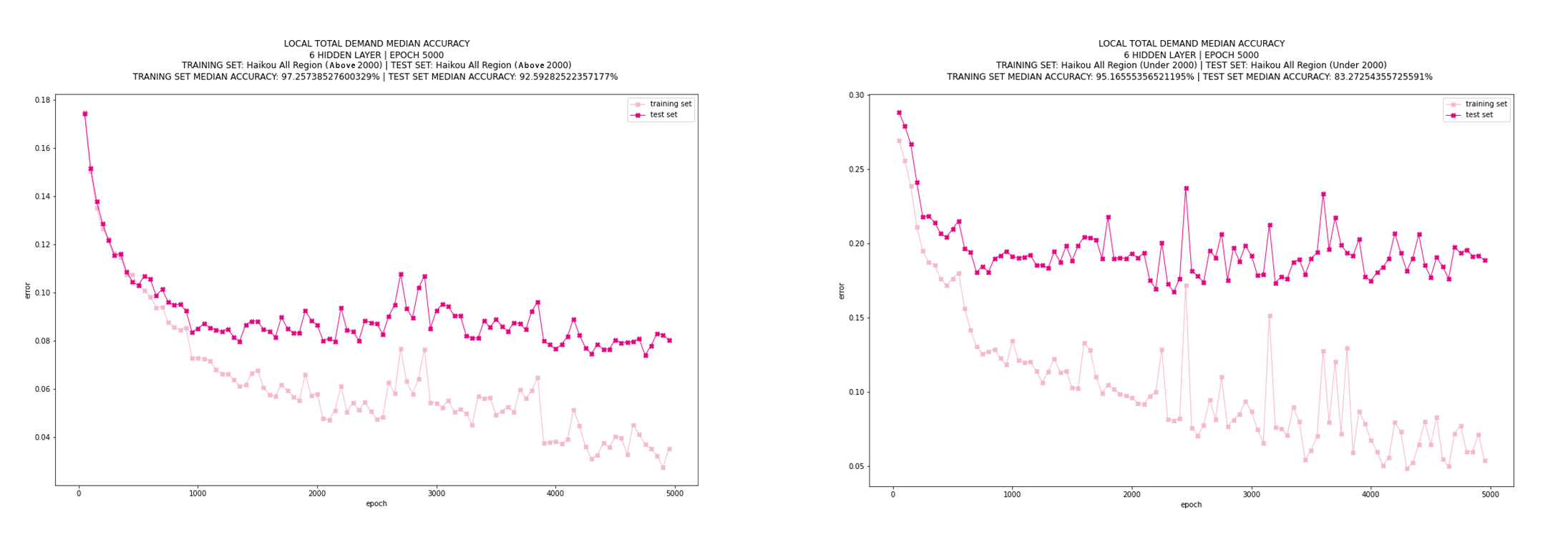}
    \caption{The prediction error rates of daily total VHT amount, Right: trained with samples having accumulated VHT of less than 2000 hours within a month, Left: trained with samples having accumulated VHT of more than 2000 hours within a month.}
    \label{compare_2000}
\end{figure}

Figure \ref{F26} and \ref{F27} present the error rates of the demand forecast models both in the urban and whole area of Haikou, showing they fall to a convergent state as the number of training iterations increases. The results plotted in the figures fulfill expectations regarding the prediction accuracy of the Haikou urban model test: 95.68\% for the average total VHT throughout one day, 97.84\% for the VHT temporal distribution prediction. The prediction accuracy of the Haikou all-region model test boasts an accuracy of 85.77\% for the average total VHT throughout one day, 95.17\% for the VHT temporal distribution prediction.\par
\begin{figure*}[!h]
    \includegraphics[width=1\textwidth]{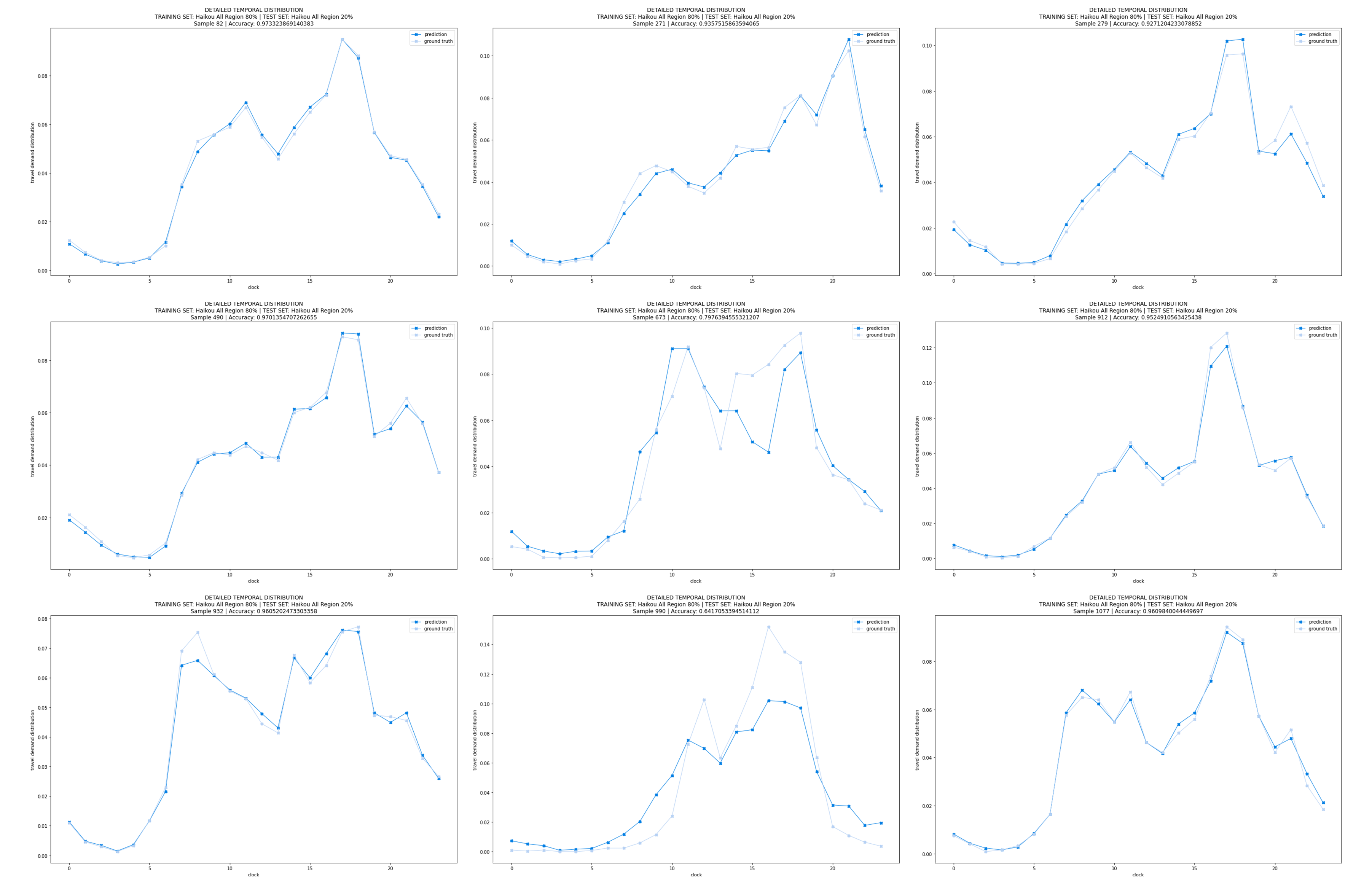}
    \caption{Instances of the prediction results (bright lines) calculated by our neural network are compared with the ground truths (faded lines). }
    \label{F28}
\end{figure*}
\begin{figure}[!h]
    \includegraphics[width=0.85\textwidth]{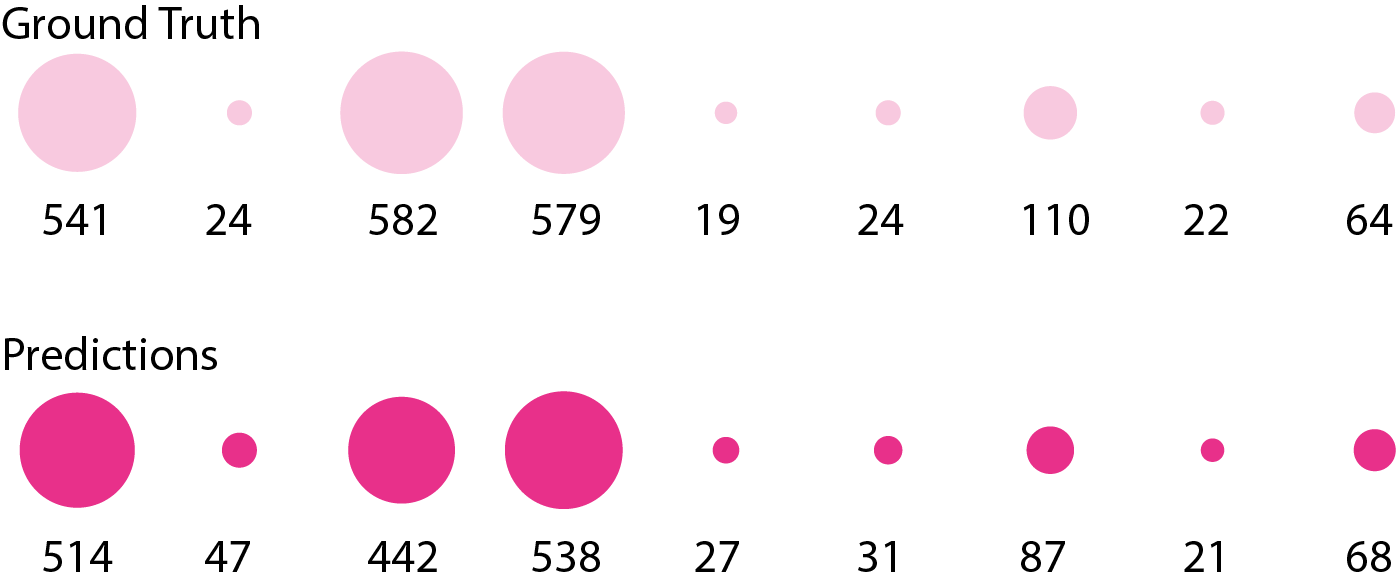}
    \caption{These circle areas represent the values of the ground truth or the prediction of daily average VHT taking place in nine samples (from left to right: ST82, ST271, ST279, ST490, ST673, ST912, ST932, ST990, ST1077).}
    \label{F29}
\end{figure}

Nine samples from the all-region model test were randomly calculated, had their prediction outputs visualized, and then compared to the ground truths. The built environment inputs and the detailed demand predictions are displayed in the table \ref{table:3} and \ref{table:4}.  The bright lines of Figure \ref{F28} present the prediction results of these instances calculated by our neural network, and the corresponding ground truths represented by the faded lines are also plotted in these figures to make a comparison (i.e., ST82, ST271, ST279, ST490, ST673, ST912, ST932, ST990, and ST1077 represent different samples from the test set, which annotate their numerical positions). Figure \ref{F29} depicts the predictions and the ground truths of total VHT throughout one day.\par
An accuracy of 85.77\% in the average total VHT prediction of Haikou is a relatively low value when compared with the results draw from exclusively its urban area, but the prediction results seem acceptable when visualized into circle areas. Most cases with low accuracy occur in remote places that have few and random travel orders compared with those located in urban areas. An understandable reason for this discrepancy is the lack of travel orders to form a consistent and regular pattern in these samples; therefore, slight differences in absolute value lead to notable error rates as ST271 shows in figure \ref{F29}. \par
To determine if samples with sparse travel activity cause the problem of accuracy, we roughly divided the all-region data set into two parts. One contains 3065 samples, each of which accumulated VHT of no more than 2000 hours within a month (Set-A), and the other part contains 2935 samples that each accrued VHT above 2000 hours (Set-B). Two computational models are therefore trained with these two data sets and preliminary network structure. Figure \ref{compare_2000} plots the median error rates of both models' predictions in daily average total VHT amount. They show that convergence and overfitting of the model trained by Set-A occurred at approximately 1000 epochs, and that accuracy sits at a relatively low at 83.27\%. After removing the data samples with low travel activity, the model trained with the retained data shows a higher accuracy rate of 92.59\% and a stabler error fitting curve. \par
The results are visualized and analyzed for VHT temporal distribution predictions as well. Our calculation model sports a significant fitting effect on the detailed temporal proportion distribution of travel demand. According to the same problem above, some samples in remote places without abundant vehicle orders to produce VHT incompletely fit the truths as ST990 shows, which has an accuracy of 64.17\% below the median accuracy. \par

In conclusion, we received satisfying median accuracy results about the neural network used to map a link between a given built environment and vehicle travel demand prediction. The detailed temporal distribution of the demand offers acceptable median accuracy as well. For total demand, despite the presence of samples with insufficient travel activities muddling their demand patterns, the magnitude of the total remains relatively correct and distinguishable from samples with significant travel activities through the visualization process. These verify the feasibility of our computational models to forecast vehicle travel demand in different periods, and these models perform better in areas with frequent human activities compared to remote ones.\par



\begin{table*}[!htb]
\begin{tabular}{|c c c c c c c c c c|}
\hline
Inputs & ST82 & ST271 & ST279 & ST490 & ST673 & ST912 & ST932 & ST990 & ST1077\\
\hline
poi 00  & 6.94\%    & 8.62\%     & 1.38\%     & 4.32\%     & 3.57\%     & 7.82\%     & 10.22\%    & 2.38\%     & 3.54\%      \\
poi 01  & 14.30\%   & 10.92\%    & 17.59\%    & 9.31\%     & 5.95\%     & 8.64\%     & 14.06\%    & 7.14\%     & 9.60\%      \\
poi 02  & 10.93\%   & 7.47\%     & 19.66\%    & 6.06\%     & 16.67\%    & 31.48\%    & 9.74\%     & 26.19\%    & 11.11\%     \\
poi 03  & 17.87\%   & 8.62\%     & 15.17\%    & 9.31\%     & 3.57\%     & 4.73\%     & 17.57\%    & 11.90\%    & 13.13\%     \\
poi 04  & 2.94\%    & 3.45\%     & 6.55\%     & 8.59\%     & 0.00\%     & 1.03\%     & 1.60\%     & 0.00\%     & 2.02\%      \\
poi 05  & 6.87\%    & 5.17\%     & 1.03\%     & 5.31\%     & 2.38\%     & 1.03\%     & 4.47\%     & 0.00\%     & 2.53\%      \\
poi 06  & 6.24\%    & 0.57\%     & 1.38\%     & 8.63\%     & 1.19\%     & 1.65\%     & 3.51\%     & 7.14\%     & 5.56\%      \\
poi 07  & 0.56\%    & 1.15\%     & 1.72\%     & 0.44\%     & 0.00\%     & 0.82\%     & 0.32\%     & 0.00\%     & 0.51\%      \\
poi 08  & 6.24\%    & 1.72\%     & 1.38\%     & 6.26\%     & 0.00\%     & 2.47\%     & 5.91\%     & 0.00\%     & 13.13\%     \\
poi 09  & 8.69\%    & 32.76\%    & 14.48\%    & 13.66\%    & 33.33\%    & 20.58\%    & 11.50\%    & 11.90\%    & 7.58\%      \\
poi 10  & 7.85\%    & 5.17\%     & 7.93\%     & 5.23\%     & 10.71\%    & 6.58\%     & 7.03\%     & 2.38\%     & 5.05\%      \\
poi 11  & 4.48\%    & 9.77\%     & 4.48\%     & 12.71\%    & 20.24\%    & 9.05\%     & 8.31\%     & 23.81\%    & 17.68\%     \\
poi 12  & 0.84\%    & 0.00\%     & 0.34\%     & 0.04\%     & 0.00\%     & 0.00\%     & 0.00\%     & 0.00\%     & 0.00\%      \\
poi 13  & 1.33\%    & 1.15\%     & 0.69\%     & 0.83\%     & 2.38\%     & 2.67\%     & 2.40\%     & 2.38\%     & 4.04\%      \\
poi 14  & 2.80\%    & 2.30\%     & 1.03\%     & 7.80\%     & 0.00\%     & 0.00\%     & 1.60\%     & 0.00\%     & 1.52\%      \\
poi 15  & 1.12\%    & 1.15\%     & 5.17\%     & 1.50\%     & 0.00\%     & 1.44\%     & 1.76\%     & 4.76\%     & 3.03\%           \\
\hline
Overall & 1427      & 174        & 290        & 2525       & 84         & 486        & 626        & 42         & 198         \\
\hline
\end{tabular}
\caption{The input table for Figure \ref{F28}.}
\label{table:3}
\end{table*}

\begin{table*}[!htb]
\begin{tabular}{|c c c c c c c c c c|}
\hline
Outputs & ST82 & ST271 & ST279 & ST490 & ST673 & ST912 & ST932 & ST990 & ST1077\\
\hline
vht 00      & 1.09\%    & 1.19\%     & 1.93\%     & 1.92\%     & 1.19\%     & 0.77\%     & 1.14\%     & 0.73\%     & 0.81\%      \\
vht 01      & 0.66\%    & 0.54\%     & 1.27\%     & 1.45\%     & 0.54\%     & 0.44\%     & 0.48\%     & 0.52\%     & 0.44\%      \\
vht 02      & 0.38\%    & 0.30\%     & 1.03\%     & 0.95\%     & 0.34\%     & 0.16\%     & 0.34\%     & 0.40\%     & 0.23\%      \\
vht 03      & 0.25\%    & 0.21\%     & 0.47\%     & 0.61\%     & 0.21\%     & 0.10\%     & 0.15\%     & 0.09\%     & 0.16\%      \\
vht 04      & 0.33\%    & 0.32\%     & 0.46\%     & 0.51\%     & 0.33\%     & 0.19\%     & 0.37\%     & 0.16\%     & 0.28\%      \\
vht 05      & 0.50\%    & 0.49\%     & 0.50\%     & 0.48\%     & 0.33\%     & 0.53\%     & 1.17\%     & 0.21\%     & 0.84\%      \\
vht 06      & 1.16\%    & 1.11\%     & 0.79\%     & 0.92\%     & 0.94\%     & 1.15\%     & 2.16\%     & 0.63\%     & 1.63\%      \\
vht 07      & 3.44\%    & 2.50\%     & 2.17\%     & 2.93\%     & 1.21\%     & 2.48\%     & 6.42\%     & 1.18\%     & 5.86\%      \\
vht 08      & 4.87\%    & 3.41\%     & 3.20\%     & 4.11\%     & 4.64\%     & 3.28\%     & 6.59\%     & 2.04\%     & 6.80\%      \\
vht 09      & 5.56\%    & 4.40\%     & 3.91\%     & 4.42\%     & 5.46\%     & 4.80\%     & 6.08\%     & 3.84\%     & 6.23\%      \\
vht 10      & 6.02\%    & 4.60\%     & 4.56\%     & 4.47\%     & 9.11\%     & 5.00\%     & 5.59\%     & 5.15\%     & 5.48\%      \\
vht 11      & 6.90\%    & 3.95\%     & 5.33\%     & 4.84\%     & 9.11\%     & 6.38\%     & 5.31\%     & 7.53\%     & 6.41\%      \\
vht 12      & 5.57\%    & 3.76\%     & 4.83\%     & 4.30\%     & 7.45\%     & 5.43\%     & 4.79\%     & 6.97\%     & 4.62\%      \\
vht 13      & 4.78\%    & 4.43\%     & 4.29\%     & 4.31\%     & 6.40\%     & 4.56\%     & 4.31\%     & 5.97\%     & 4.17\%      \\
vht 14      & 5.86\%    & 5.27\%     & 6.10\%     & 6.13\%     & 6.41\%     & 5.16\%     & 6.67\%     & 8.08\%     & 5.39\%      \\
vht 15      & 6.71\%    & 5.51\%     & 6.36\%     & 6.16\%     & 5.07\%     & 5.53\%     & 6.00\%     & 8.23\%     & 5.86\%      \\
vht 16      & 7.23\%    & 5.48\%     & 7.00\%     & 6.56\%     & 4.62\%     & 10.95\%    & 6.82\%     & 10.19\%    & 7.18\%      \\
vht 17      & 9.49\%    & 6.88\%     & 10.18\%    & 9.04\%     & 8.19\%     & 12.09\%    & 7.62\%     & 10.12\%    & 9.21\%      \\
vht 18      & 8.73\%    & 8.10\%     & 10.26\%    & 9.00\%     & 8.92\%     & 8.67\%     & 7.56\%     & 9.70\%     & 8.75\%      \\
vht 19      & 5.67\%    & 7.19\%     & 5.36\%     & 5.18\%     & 5.58\%     & 5.29\%     & 4.82\%     & 5.41\%     & 5.72\%      \\
vht 20      & 4.64\%    & 9.05\%     & 5.25\%     & 5.39\%     & 4.04\%     & 5.57\%     & 4.49\%     & 3.14\%     & 4.44\%      \\
vht 21      & 4.52\%    & 10.78\%    & 6.12\%     & 6.26\%     & 3.43\%     & 5.76\%     & 4.82\%     & 3.08\%     & 4.79\%      \\
vht 22      & 3.45\%    & 6.49\%     & 4.85\%     & 5.64\%     & 2.91\%     & 3.61\%     & 3.38\%     & 1.77\%     & 3.33\%      \\
vht 23      & 2.20\%    & 3.82\%     & 3.39\%     & 3.74\%     & 2.09\%     & 1.84\%     & 2.60\%     & 1.95\%     & 2.12\%         \\
\hline
Overall & 541.36 & 23.84 & 582.38 & 579.24 & 18.99 & 23.66 & 109.62 & 22.33 & 64.20           \\
\hline
\end{tabular}
\caption{The forecast table for Figure \ref{F28}.}
\label{table:4}
\end{table*}

\subsection{Cross-city Experiment}

With the aid of trained computational models, we successfully analyzed and judged the results of traffic demand in Haikou based on the given built environment information. However, it remains to be verified if this prediction model can be migrated across different regions. The twofold issue manifests as such: (1) whether a predictive model trained in one city can be applied to another city, and (2) whether a model trained in one part of a city can be used for the entire area. To verify the universality of the relationship between environment and vehicle travel demand and further demonstrate the applicability of our method, we trained the neural networks for different areas (Figure \ref{F30}) and conducted a hybrid test. In addition, although ANNs performed best in fitting the mapping relationship in the preliminary five-fold accuracy verification test, algorithms such as RF and Linear SVR are also reconsidered while accounting for the modification in training data and test sets. \par
This section details the final accuracy of our prediction models in the experiments involving both the urban and whole of Haikou, Chengdu urban area, and algorithmic adjustments for better performance in multi-area applications. Data sampling and model training processes were conducted on these different areas, and the trained models are applied to the other two regions to acquire accuracy. By comparing and analyzing the median accuracy of models applied to different areas and graphing the prediction results, we determine the extent to which the prediction model can be reused. \par

\subsubsection{Aggregate Forecasting: Region Migration}

\begin{figure*}[!htb]
    \includegraphics[width=1\textwidth]{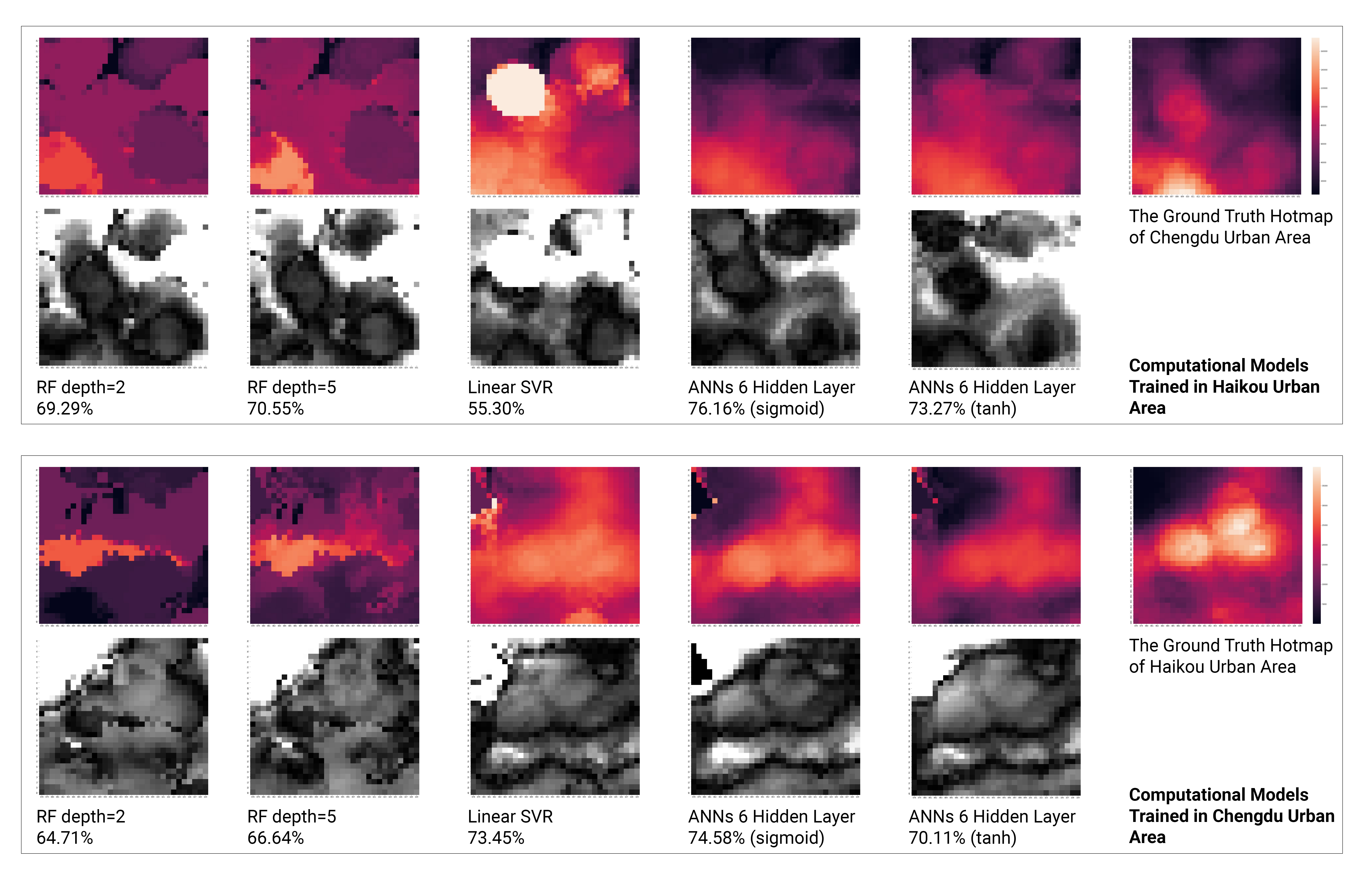}
    \caption{Accuracy and visualization of total vehicle travel demand in the cross-city experiments.}
    \label{predictions_layout}
\end{figure*}

First, on the topic of VHT aggregate forecasts used across regions, we applied the model trained to the center of one city and then to that of another. Secondly, the prediction values are visualized to show the effect of the computational model and compare the experimental results of different algorithms and datasets (Figure \ref{predictions_layout}). The predicted total VHT for each sample is remapped to the value before normalization and translated into a color value. Brighter and warmer colors represent higher values. Sample color blocks are reassembled in the area of interest and measured onto a map in km, depicting the predicted total vehicle travel demand as a macroscopic geographic distribution to be compared to reality. In addition, the relative error between the predicted and true values for each sample is visualized as a color value. Blacker means a higher accuracy of the prediction model whereas whiter represents a higher error value. Similarly, the error value color block is reallocated onto the map.\par

One model is trained in Haikou, then tested in Chengdu, and vice versa with the other (Figure \ref{predictions_layout}). After comparing the prediction images generated by the different algorithms and the median accuracy of the samples, we suggest that ANNs bear the best prediction results. ANNs boast not only higher median accuracy but also generate images with prediction values closest to the real VHT distribution. The prediction results of RF have a low subdivision in the color value of VHT aggregate. The degree of subdivision and accuracy increase with the tree depth value, but the change becomes extremely subtle when depth exceeds 5. The accuracy of SVM is also not as high as ANNs, and according to the visualization of the predicted values, SVM overestimates the VHT value of samples with sparse travel activity. \par
The visualization of VHT generated by ANNs matches the spatial distribution trend of the real situation. To clarify, total VHT in Chengdu decreases from the lower-left outward in the predicted image, and the total VHT in Haikou aggregates laterally in the middle of the predicted map. This indicates that the prediction model for the total VHT amount works for another city whose data is completely unlearned by the artificial neural network during the training process.\par
However, the color value distribution tends toward average more than the real. We tested different activation functions to obtain better prediction results. The experiments show that Relu and Softmax fail to fit the training of the prediction model. The results of Tanh and Sigmoid are similar. We ultimately chose Sigmoid as the final activation function because it provides a more even distribution of color values while Tanh generates predictions with slightly lower accuracy. Under those conditions, Sigmoid brightens the darker parts of the upper left and lower right diagonal of Chengdu prediction. In the Haikou prediction image, it darkens the central area (Figure \ref{predictions_layout}).\par
Additionally, the accuracy images of ANNs show the sample with ground truths of higher total VHT amount have darker color values (i.e., the ANNs model holds a higher prediction accuracy on that sample). However, the sample with the highest accuracy is not at the center of the city, where the highest VHT values converge, but in the outer ring, where the total VHT values are slightly lower. The black area in the accuracy plot consistently appears as a ring, and the interior contains dark gray blocks slightly brighter compared to the edges, meaning the accuracy of the ANN model first starts higher and lowers from the center of the city outward.\par
In general, the prediction models for VHT aggregate are migratable between different urban areas provided that ANNs use Sigmoid, possessing a median accuracy of around 75\%, increasing and then decreasing from the city center outwards.

\subsubsection{Aggregate Forecasting: Region Expansion}
However, there exists the question of applying models trained in a small-scale urban area dataset to a larger scale one. While we have achieved excellent accuracy in the entire Haikou area in our previous accuracy experiments using ANNs in a randomly distributed geographic sample, cross-regional experiments using ANNs, RF, and SVM show extremely low accuracy when the models used for Haikou are applied to a larger-scale citywide area (Table \ref{table:5}). This is due to certain randomness in areas with sparse travel activity, a conclusion strengthened by the accuracy visualizations (Figure \ref{predictions_layout}). Samples with high error rates, marked with bright colors, in the accuracy images are always sparse in total VHT, which are shown as dark color in the ground truth images. As previously stated in the accuracy experiments, it is difficult for the algorithms to acquire irregular patterns from the data samples with sparse travel activity. Valid mapping relationships in these remote areas acquired by the algorithms in areas with dense travel activity are also a challenge. In applying the model to an entire city area, the algorithm overfitting occurs when exclusively trained on the urban center region. \par

\begin{table}[!h]
\begin{tabular}{|l|l|l|}
\hline
\multicolumn{2}{|l|}{Different computational models} & Accuracy \\ \hline
\multirow{2}{*}{ANNs}             & Haikou Urban     & 17.21\%      \\
                                  & Chengdu Urban    & 33.35\%      \\ \hline
\multirow{2}{*}{Linear SVR}       & Haikou Urban     & -65.69\%     \\
                                  & Chengdu Urban    & -21.49\%     \\ \hline
\multirow{2}{*}{Random Forest}    & Haikou Urban     & -178.32\%    \\
                                  & Chengdu Urban    & -269.16\%    \\ \hline
\end{tabular}
\caption{This table records the accuracy of VHT amount predictions in entire Haikou area.}
\label{table:5}
\end{table}

\begin{table}[!h]
\begin{tabular}{|l|l|l|}
\hline
              & Model U→Set A & Model A→Set U \\ \hline
ANNs          & 36.10\%       & 61.67\%       \\
Random Forest & 30.12\%       & 52.00\%       \\
Linear SVR    & -65.68\%      & 15.91\%       \\ \hline
\end{tabular}
\caption{The all-region data set is divided into two parts, one of which  (A)  has monthly VHT accumulation above 2000 hours on each sample, the other (U) has monthly VHT accumulation under 2000 hours on each sample. This table shows the prediction accuracy of three algorithms between the two data sets.}
\label{table:6}
\end{table}

\begin{figure}[]
    \includegraphics[width=0.8\textwidth]{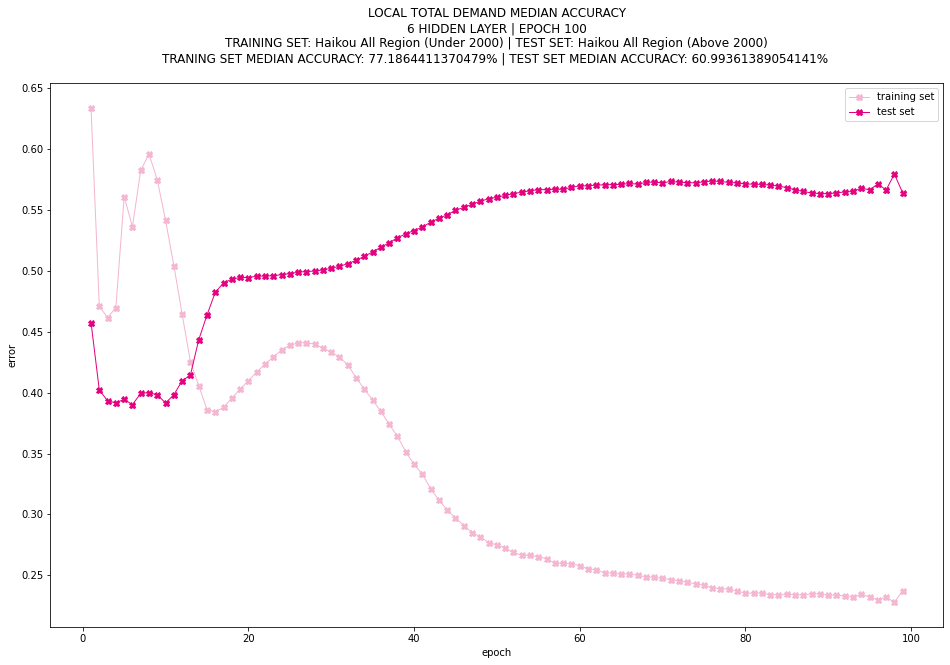}
    \caption{Over-fitting of ANNs in the prediction of the total VHT amount.}
    \label{hkaa_hkau}
\end{figure}

\begin{table*}[!htb]
\begin{tabular}{|l|l|l|l|l|}
\hline
\multicolumn{2}{|l|}{}                                                   & \multicolumn{3}{l|}{Accuracy of temporal distribution in different test areas} \\ \hline
\multicolumn{2}{|l|}{Different computational models}                     & Haikou All-region     & Haikou Urban    & Chengdu Urban    \\ \hline
\rowcolor[HTML]{EFEFEF} 
\cellcolor[HTML]{EFEFEF}                             & Haikou All-region & -                     & 93.58\%           & 80.04\%            \\
\rowcolor[HTML]{EFEFEF} 
\cellcolor[HTML]{EFEFEF}                             & Haikou Urban      & 83.47\%                  & -               & 77.28\%             \\
\rowcolor[HTML]{EFEFEF} 
\multirow{-3}{*}{\cellcolor[HTML]{EFEFEF}ANNs}       & Chengdu Urban     & 77.05\%                  & 80.43\%            & -                \\ \hline
                                                     & Haikou All-region & -                     & 89.31\%            & 79.02\%             \\
                                                     & Haikou Urban      & 82.95\%                  & -               & 75.79\%             \\
\multirow{-3}{*}{Random Forest}                      & Chengdu Urban     & 76.93\%                  & 78.19\%            & -                \\ \hline
\rowcolor[HTML]{EFEFEF} 
\cellcolor[HTML]{EFEFEF}                             & Haikou All-region & -                     & 91.34\%            & 78.22\%             \\
\rowcolor[HTML]{EFEFEF} 
\cellcolor[HTML]{EFEFEF}                             & Haikou Urban      & 80.38\%                  & -               & 73.74\%             \\
\rowcolor[HTML]{EFEFEF} 
\multirow{-3}{*}{\cellcolor[HTML]{EFEFEF}Linear SVR} & Chengdu Urban     & 67.68\%                  & 76.50\%            & -                \\ \hline
\end{tabular}
\caption{This table records the accuracy results of temporal distribution in the cross-area experiment.}
\label{table:7}
\end{table*}

\begin{figure*}[!htb]
    \includegraphics[width=1\textwidth]{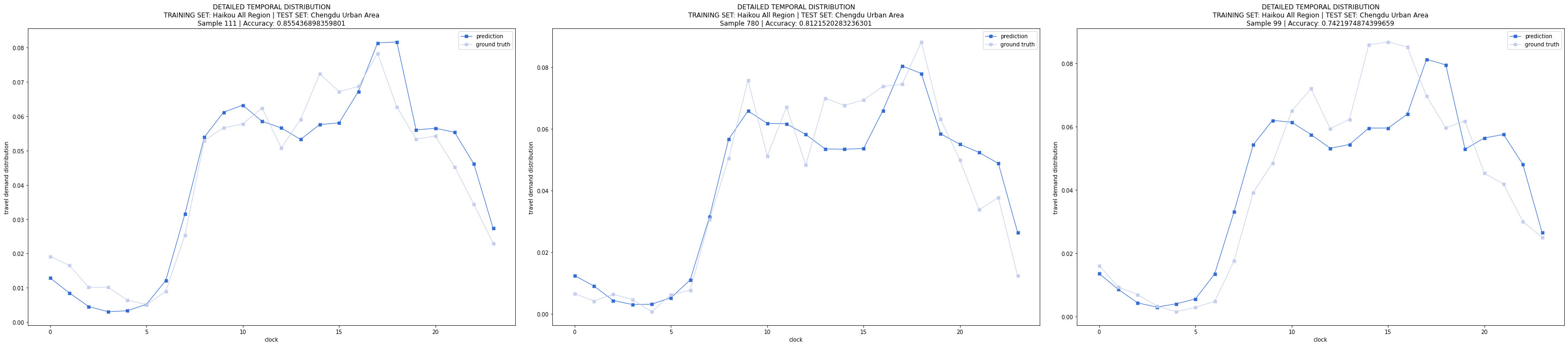}
    \caption{The output instances of prediction model which uses the same training set as figure \ref{F28} but is tested in Chengdu}
    \label{ST_hka_cdu}
\end{figure*}

Another experiment is conducted to illustrate the over-fitting phenomenon that transpires in citywide areas. The all-region dataset once again is divided into two parts. One contains monthly VHT accumulation above 2000 hours on each sample. The other has a monthly VHT accumulation under 2000 hours. Using the multi-area test method, we tested the capability of three algorithms trained on one data set and then be applied on the other (Table \ref{table:6}). In reference to ANNs, the model of artificial neural network over-fits in the early stage of training according to the error rate curve (Figure \ref{hkaa_hkau}), leading to the model of ANNs having an extremely low accuracy. \par

\subsubsection{Temporal Distribution Forecasting}

For a detailed VHT temporal distribution, the median accuracy values obtained in the experiments closely approach the expected (Table \ref{table:7}) in both test region expansion and migration. The prediction accuracy of ANNs remains higher than that of SVM and RF. This is true in all the multi-area tests. According to the table \ref{table:7}, taking the multi-area test between urban areas of Chengdu and Haikou for example, the median accuracy reaches about 80\% when the computational model is applied to a broader urban scale or the built environment of another urban area. It means that the prediction model has applicability among different urban environments, and the mapping relation acquired in one area can be reused elsewhere to a certain degree. To better demonstrate the accuracy of the prediction model's versatility, we selected and visualized three samples' temporal distribution predictions using all of Haikou as the training set and Chengdu environmental data as the input of the test process. The predictions of these three samples have similar patterns and the accuracy of them exceeds (85.54\%), hovers around (81.22\%), or strays below the median (74.22\%) Figure \ref{ST_hka_cdu}). Compared with the richer patterns in figure \ref{F28}, the model predicts a time distribution on these Chengdu samples close to real-time situations.\par

We have verified a link between our living environment and vehicle travel demand, which is consistent with previous studies of Cervero et al., and attempted to apply this relationship learned by the computational model in one area to a broader urban or other areas. Although there exist issues in predicting the total amount of VHT when applying computational models from congested city zones to fringes with scant activity, migration experiments have demonstrated the potential of applying models trained in one area to others. Thus, using this approach, the dataset from a limited number of areas can be utilized to map the relationship between living environment and vehicle travel demand. This will help urban researchers and designers in predicting detailed travel demand and offer feedback on improvements in the living environment, thus significantly reducing their upfront research costs.\par


\subsection{AI-assisted Urban Design}
In addition, we experimented with the final model results in terms of design applications. The final hybrid prediction model consists of a combination of ANNs with six and seven hidden layers for predicting, respectively, the amount of vehicle travel demand and the related temporal distribution. As stated previously, a large body of prior research showed that mixed land use can promote non-vehicular traveling and thus reduce demand. However, the extent and specific effects of functional mixing are vague in most studies. Without such references, designers struggle to intuitively and quickly obtain feedback on functional blends. To demonstrate the value of our approach in this scenario, hybrid computational models are used in this study to explore a novel design approach.\par

\begin{figure*}[!htb]
    \includegraphics[width=1\textwidth]{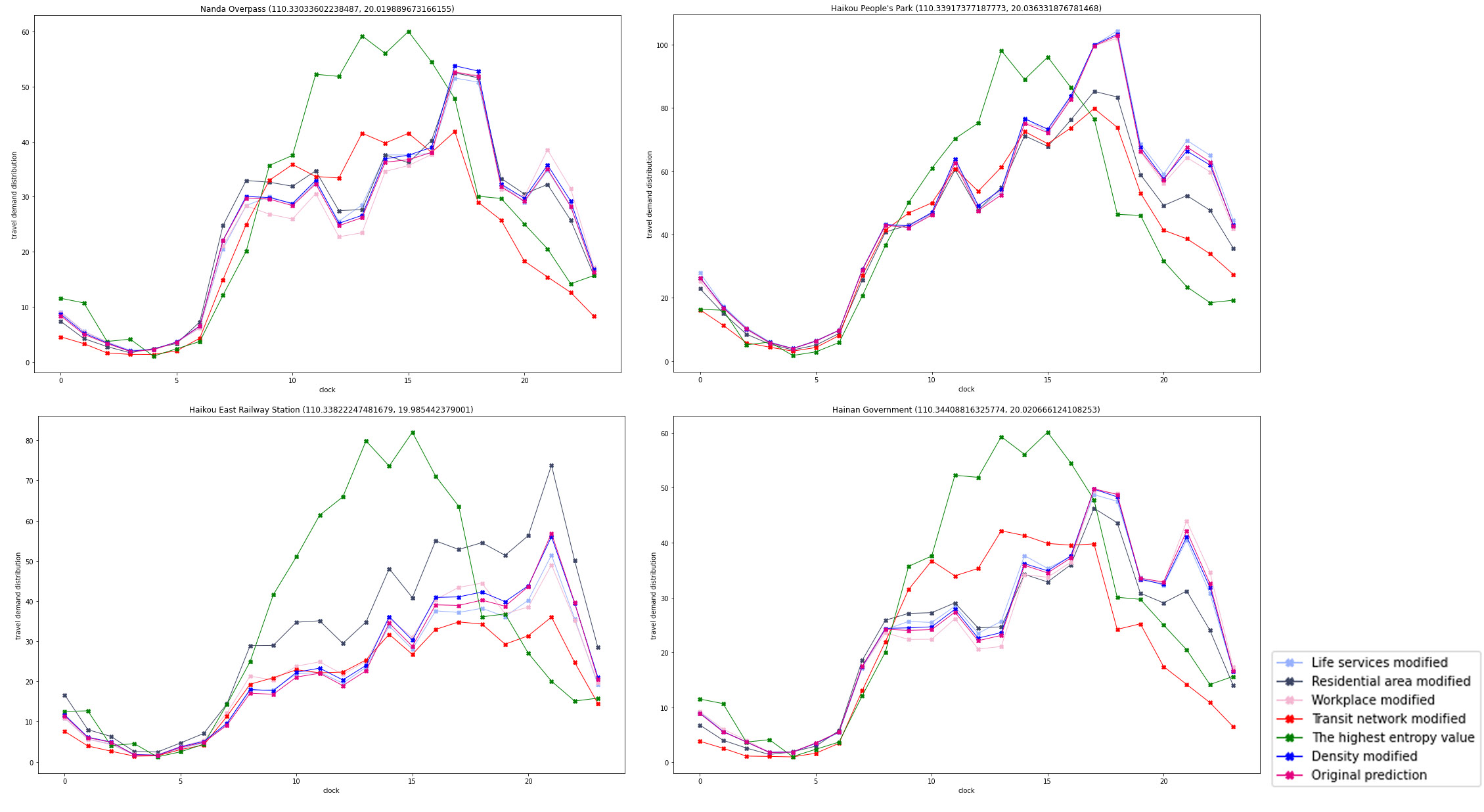}
    \caption{The output instances of prediction model which uses the same training set as figure \ref{F28} but is tested in Chengdu}
    \label{modified_predictions}
\end{figure*}
\begin{figure}[!htb]
    \includegraphics[width=1\textwidth]{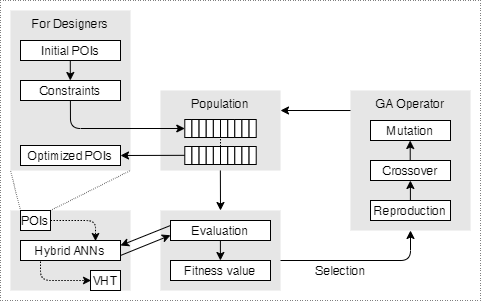}
    \caption{the combination of genetic algorithms and hybrid ANNs}
    \label{GA}
\end{figure}

We selected and adjusted the urban functions of the buffer areas near four sample points: Nanda Overpass (110.3303, 20.0199), Haikou People's Park (110.3392, 20.0363), Haikou East Railway Station (110.3382, 19.9854), and Hainan Government (110.3441, 20.0207). Four different types of urban functions are focused in the study related to the choice of travel patterns: restaurants, residential areas, workplaces, and public transportation. In addition, we endeavored to predict the case where the original function ratios are maintained while the density thereof increases, including where the number of urban functions is the same for all types (i.e., the case where the entropy of the urban function mix is highest). The table \ref{table:8} records our inputs for seven different functional ratios, including the original functional numbers, at four locations. Simultaneously, the figure \ref{modified_predictions} records the outputs of our hybrid computational model for predicting vehicle travel demand in these modified living environments. Notably, the designer must interpret the results as associative because the link between living environment and travel demand learned by the computational model is not causal in nature.\par
The conclusion that can be drawn from the figure \ref{modified_predictions} is that a perfectly average distribution of functions does not imply an optimal distribution of traffic demand. It is a possibility that the largest entropy in a mix of urban functions begets an even more irrational distribution of traffic demand. The green line in the figure represents numerical uniformity in all the functions, and peak traffic demand not only stagnates but indeed could increase. This again illustrates the potential for misleading conclusions to be drawn from vaguely describing information in the urban environment. Conversely, among the adjustment measures for the four explicit functional categories, the approach with a clear effect involves mainly public transport. A reduction of traffic pressure in all four locations can be notably observed when public transport stations increase. At the same time, an increase of residential functions for the peak traffic periods depends on the situation. Adding residential block modules near Haikou People's Park corresponds to the relief of traffic pressure. Similar measures performed near East Station instead correspond to higher traffic demand.\par

\begin{table*}[!htb]
\begin{tabular}{llllllllllllllllll}
                          & \multicolumn{17}{c}{Nanda Overpass}                                                    \\ \hline
\multicolumn{1}{|l|}{POI} & 00                         & 01                          & 02                          & 03                          & 04                          & 05                          & 06                          & 07                         & 08                          & 09                          & 10                          & 11                          & 12                        & 13                          & 14                          & 15                          & \multicolumn{1}{l|}{sum}  \\ \hline
\multicolumn{1}{|l|}{1}   & \cellcolor[HTML]{F2F2F2}88 & \cellcolor[HTML]{F2F2F2}19  & \cellcolor[HTML]{F2F2F2}10  & \cellcolor[HTML]{F2F2F2}18  & \cellcolor[HTML]{F2F2F2}72  & \cellcolor[HTML]{F2F2F2}103 & \cellcolor[HTML]{F2F2F2}112 & \cellcolor[HTML]{F2F2F2}3  & \cellcolor[HTML]{F2F2F2}122 & \cellcolor[HTML]{F2F2F2}44  & \cellcolor[HTML]{F2F2F2}108 & \cellcolor[HTML]{F2F2F2}71  & \cellcolor[HTML]{F2F2F2}0 & \cellcolor[HTML]{F2F2F2}27  & \cellcolor[HTML]{F2F2F2}90  & \cellcolor[HTML]{F2F2F2}16  & \multicolumn{1}{l|}{904}  \\
\multicolumn{1}{|l|}{2}   & 88                         & \cellcolor[HTML]{DDEBF7}35  & \cellcolor[HTML]{DDEBF7}26  & \cellcolor[HTML]{DDEBF7}34  & \cellcolor[HTML]{DDEBF7}88  & \cellcolor[HTML]{DDEBF7}119 & 112                         & 3                          & 122                         & 44                          & 108                         & 71                          & 0                         & 27                          & 90                          & 16                          & \multicolumn{1}{l|}{984}  \\
\multicolumn{1}{|l|}{3}   & 88                         & 19                          & 10                          & 18                          & 72                          & 103                         & 112                         & 3                          & \cellcolor[HTML]{DDEBF7}202 & 44                          & 108                         & 71                          & 0                         & 27                          & 90                          & 16                          & \multicolumn{1}{l|}{984}  \\
\multicolumn{1}{|l|}{4}   & 88                         & 19                          & 10                          & 18                          & 72                          & 103                         & 112                         & 3                          & 122                         & \cellcolor[HTML]{DDEBF7}84  & \cellcolor[HTML]{DDEBF7}128 & \cellcolor[HTML]{DDEBF7}91  & 0                         & 27                          & 90                          & 16                          & \multicolumn{1}{l|}{984}  \\
\multicolumn{1}{|l|}{5}   & 88                         & 19                          & 10                          & 18                          & 72                          & 103                         & 112                         & 3                          & 122                         & 44                          & 108                         & 71                          & 0                         & \cellcolor[HTML]{DDEBF7}107 & 90                          & 16                          & \multicolumn{1}{l|}{984}  \\
\multicolumn{1}{|l|}{6}   & \cellcolor[HTML]{DDEBF7}-  & \cellcolor[HTML]{DDEBF7}-   & \cellcolor[HTML]{DDEBF7}-   & \cellcolor[HTML]{DDEBF7}-   & \cellcolor[HTML]{DDEBF7}-   & \cellcolor[HTML]{DDEBF7}-   & \cellcolor[HTML]{DDEBF7}-   & \cellcolor[HTML]{DDEBF7}-  & \cellcolor[HTML]{DDEBF7}-   & \cellcolor[HTML]{DDEBF7}-   & \cellcolor[HTML]{DDEBF7}-   & \cellcolor[HTML]{DDEBF7}-   & \cellcolor[HTML]{DDEBF7}- & \cellcolor[HTML]{DDEBF7}-   & \cellcolor[HTML]{DDEBF7}-   & \cellcolor[HTML]{DDEBF7}-   & \multicolumn{1}{l|}{984}  \\
\multicolumn{1}{|l|}{7}   & \cellcolor[HTML]{DDEBF7}95 & \cellcolor[HTML]{DDEBF7}20  & \cellcolor[HTML]{DDEBF7}10  & \cellcolor[HTML]{DDEBF7}19  & \cellcolor[HTML]{DDEBF7}78  & \cellcolor[HTML]{DDEBF7}112 & \cellcolor[HTML]{DDEBF7}121 & \cellcolor[HTML]{DDEBF7}3  & \cellcolor[HTML]{DDEBF7}133 & \cellcolor[HTML]{DDEBF7}47  & \cellcolor[HTML]{DDEBF7}117 & \cellcolor[HTML]{DDEBF7}77  & \cellcolor[HTML]{DDEBF7}0 & \cellcolor[HTML]{DDEBF7}29  & \cellcolor[HTML]{DDEBF7}97  & \cellcolor[HTML]{DDEBF7}17  & \multicolumn{1}{l|}{984}  \\ \hline
                          & \multicolumn{17}{c}{Haikou People's Park}                                                                                                                                             \\ \hline
\multicolumn{1}{|l|}{POI} & 00                         & 01                          & 02                          & 03                          & 04                          & 05                          & 06                          & 07                         & 08                          & 09                          & 10                          & 11                          & 12                        & 13                          & 14                          & 15                          & \multicolumn{1}{l|}{sum}  \\ \hline
\multicolumn{1}{|l|}{1}   & \cellcolor[HTML]{F2F2F2}40 & \cellcolor[HTML]{F2F2F2}191 & \cellcolor[HTML]{F2F2F2}86  & \cellcolor[HTML]{F2F2F2}178 & \cellcolor[HTML]{F2F2F2}168 & \cellcolor[HTML]{F2F2F2}247 & \cellcolor[HTML]{F2F2F2}214 & \cellcolor[HTML]{F2F2F2}32 & \cellcolor[HTML]{F2F2F2}205 & \cellcolor[HTML]{F2F2F2}259 & \cellcolor[HTML]{F2F2F2}254 & \cellcolor[HTML]{F2F2F2}284 & \cellcolor[HTML]{F2F2F2}0 & \cellcolor[HTML]{F2F2F2}27  & \cellcolor[HTML]{F2F2F2}118 & \cellcolor[HTML]{F2F2F2}102 & \multicolumn{1}{l|}{2405} \\
\multicolumn{1}{|l|}{2}   & 40                         & \cellcolor[HTML]{DDEBF7}207 & \cellcolor[HTML]{DDEBF7}102 & \cellcolor[HTML]{DDEBF7}194 & \cellcolor[HTML]{DDEBF7}184 & \cellcolor[HTML]{DDEBF7}263 & 214                         & 32                         & 205                         & 259                         & 254                         & 284                         & 0                         & 27                          & 118                         & 102                         & \multicolumn{1}{l|}{2485} \\
\multicolumn{1}{|l|}{3}   & 40                         & 191                         & 86                          & 178                         & 168                         & 247                         & 214                         & 32                         & \cellcolor[HTML]{DDEBF7}285 & 259                         & 254                         & 284                         & 0                         & 27                          & 118                         & 102                         & \multicolumn{1}{l|}{2485} \\
\multicolumn{1}{|l|}{4}   & 40                         & 191                         & 86                          & 178                         & 168                         & 247                         & 214                         & 32                         & 205                         & \cellcolor[HTML]{DDEBF7}299 & \cellcolor[HTML]{DDEBF7}274 & \cellcolor[HTML]{DDEBF7}304 & 0                         & 27                          & 118                         & 102                         & \multicolumn{1}{l|}{2485} \\
\multicolumn{1}{|l|}{5}   & 40                         & 191                         & 86                          & 178                         & 168                         & 247                         & 214                         & 32                         & 205                         & 259                         & 254                         & 284                         & 0                         & \cellcolor[HTML]{DDEBF7}107 & 118                         & 102                         & \multicolumn{1}{l|}{2485} \\
\multicolumn{1}{|l|}{6}   & \cellcolor[HTML]{DDEBF7}-  & \cellcolor[HTML]{DDEBF7}-   & \cellcolor[HTML]{DDEBF7}-   & \cellcolor[HTML]{DDEBF7}-   & \cellcolor[HTML]{DDEBF7}-   & \cellcolor[HTML]{DDEBF7}-   & \cellcolor[HTML]{DDEBF7}-   & \cellcolor[HTML]{DDEBF7}-  & \cellcolor[HTML]{DDEBF7}-   & \cellcolor[HTML]{DDEBF7}-   & \cellcolor[HTML]{DDEBF7}-   & \cellcolor[HTML]{DDEBF7}-   & \cellcolor[HTML]{DDEBF7}- & \cellcolor[HTML]{DDEBF7}-   & \cellcolor[HTML]{DDEBF7}-   & \cellcolor[HTML]{DDEBF7}-   & \multicolumn{1}{l|}{2485} \\
\multicolumn{1}{|l|}{7}   & \cellcolor[HTML]{DDEBF7}41 & \cellcolor[HTML]{DDEBF7}197 & \cellcolor[HTML]{DDEBF7}88  & \cellcolor[HTML]{DDEBF7}183 & \cellcolor[HTML]{DDEBF7}173 & \cellcolor[HTML]{DDEBF7}255 & \cellcolor[HTML]{DDEBF7}221 & \cellcolor[HTML]{DDEBF7}33 & \cellcolor[HTML]{DDEBF7}211 & \cellcolor[HTML]{DDEBF7}267 & \cellcolor[HTML]{DDEBF7}262 & \cellcolor[HTML]{DDEBF7}293 & \cellcolor[HTML]{DDEBF7}0 & \cellcolor[HTML]{DDEBF7}27  & \cellcolor[HTML]{DDEBF7}121 & \cellcolor[HTML]{DDEBF7}105 & \multicolumn{1}{l|}{2485} \\ \hline
                          & \multicolumn{17}{c}{Haikou East Railway   Station}                                                                                                                                                                                                             \\ \hline
\multicolumn{1}{|l|}{POI} & 00                         & 01                          & 02                          & 03                          & 04                          & 05                          & 06                          & 07                         & 08                          & 09                          & 10                          & 11                          & 12                        & 13                          & 14                          & 15                          & \multicolumn{1}{l|}{sum}  \\ \hline
\multicolumn{1}{|l|}{1}   & \cellcolor[HTML]{F2F2F2}82 & \cellcolor[HTML]{F2F2F2}498 & \cellcolor[HTML]{F2F2F2}157 & \cellcolor[HTML]{F2F2F2}488 & \cellcolor[HTML]{F2F2F2}50  & \cellcolor[HTML]{F2F2F2}74  & \cellcolor[HTML]{F2F2F2}106 & \cellcolor[HTML]{F2F2F2}5  & \cellcolor[HTML]{F2F2F2}106 & \cellcolor[HTML]{F2F2F2}102 & \cellcolor[HTML]{F2F2F2}41  & \cellcolor[HTML]{F2F2F2}114 & \cellcolor[HTML]{F2F2F2}2 & \cellcolor[HTML]{F2F2F2}15  & \cellcolor[HTML]{F2F2F2}23  & \cellcolor[HTML]{F2F2F2}13  & \multicolumn{1}{l|}{1876} \\
\multicolumn{1}{|l|}{2}   & 82                         & \cellcolor[HTML]{DDEBF7}514 & \cellcolor[HTML]{DDEBF7}173 & \cellcolor[HTML]{DDEBF7}504 & \cellcolor[HTML]{DDEBF7}66  & \cellcolor[HTML]{DDEBF7}90  & 106                         & 5                          & 106                         & 102                         & 41                          & 114                         & 2                         & 15                          & 23                          & 13                          & \multicolumn{1}{l|}{1956} \\
\multicolumn{1}{|l|}{3}   & 82                         & 498                         & 157                         & 488                         & 50                          & 74                          & 106                         & 5                          & \cellcolor[HTML]{DDEBF7}186 & 102                         & 41                          & 114                         & 2                         & 15                          & 23                          & 13                          & \multicolumn{1}{l|}{1956} \\
\multicolumn{1}{|l|}{4}   & 82                         & 498                         & 157                         & 488                         & 50                          & 74                          & 106                         & 5                          & 106                         & \cellcolor[HTML]{DDEBF7}142 & \cellcolor[HTML]{DDEBF7}61  & \cellcolor[HTML]{DDEBF7}134 & 2                         & 15                          & 23                          & 13                          & \multicolumn{1}{l|}{1956} \\
\multicolumn{1}{|l|}{5}   & 82                         & 498                         & 157                         & 488                         & 50                          & 74                          & 106                         & 5                          & 106                         & 102                         & 41                          & 114                         & 2                         & \cellcolor[HTML]{DDEBF7}95  & 23                          & 13                          & \multicolumn{1}{l|}{1956} \\
\multicolumn{1}{|l|}{6}   & \cellcolor[HTML]{DDEBF7}-  & \cellcolor[HTML]{DDEBF7}-   & \cellcolor[HTML]{DDEBF7}-   & \cellcolor[HTML]{DDEBF7}-   & \cellcolor[HTML]{DDEBF7}-   & \cellcolor[HTML]{DDEBF7}-   & \cellcolor[HTML]{DDEBF7}-   & \cellcolor[HTML]{DDEBF7}-  & \cellcolor[HTML]{DDEBF7}-   & \cellcolor[HTML]{DDEBF7}-   & \cellcolor[HTML]{DDEBF7}-   & \cellcolor[HTML]{DDEBF7}-   & \cellcolor[HTML]{DDEBF7}- & \cellcolor[HTML]{DDEBF7}-   & \cellcolor[HTML]{DDEBF7}-   & \cellcolor[HTML]{DDEBF7}-   & \multicolumn{1}{l|}{1956} \\
\multicolumn{1}{|l|}{7}   & \cellcolor[HTML]{DDEBF7}85 & \cellcolor[HTML]{DDEBF7}519 & \cellcolor[HTML]{DDEBF7}163 & \cellcolor[HTML]{DDEBF7}508 & \cellcolor[HTML]{DDEBF7}52  & \cellcolor[HTML]{DDEBF7}77  & \cellcolor[HTML]{DDEBF7}110 & \cellcolor[HTML]{DDEBF7}5  & \cellcolor[HTML]{DDEBF7}110 & \cellcolor[HTML]{DDEBF7}106 & \cellcolor[HTML]{DDEBF7}42  & \cellcolor[HTML]{DDEBF7}118 & \cellcolor[HTML]{DDEBF7}2 & \cellcolor[HTML]{DDEBF7}15  & \cellcolor[HTML]{DDEBF7}23  & \cellcolor[HTML]{DDEBF7}13  & \multicolumn{1}{l|}{1956} \\ \hline
                          & \multicolumn{17}{c}{Hainan Government}                                                                                                                                        \\ \hline
\multicolumn{1}{|l|}{POI} & 00                         & 01                          & 02                          & 03                          & 04                          & 05                          & 06                          & 07                         & 08                          & 09                          & 10                          & 11                          & 12                        & 13                          & 14                          & 15                          & \multicolumn{1}{l|}{sum}  \\ \hline
\multicolumn{1}{|l|}{1}   & \cellcolor[HTML]{F2F2F2}50 & \cellcolor[HTML]{F2F2F2}16  & \cellcolor[HTML]{F2F2F2}21  & \cellcolor[HTML]{F2F2F2}50  & \cellcolor[HTML]{F2F2F2}65  & \cellcolor[HTML]{F2F2F2}66  & \cellcolor[HTML]{F2F2F2}87  & \cellcolor[HTML]{F2F2F2}8  & \cellcolor[HTML]{F2F2F2}87  & \cellcolor[HTML]{F2F2F2}166 & \cellcolor[HTML]{F2F2F2}96  & \cellcolor[HTML]{F2F2F2}75  & \cellcolor[HTML]{F2F2F2}0 & \cellcolor[HTML]{F2F2F2}15  & \cellcolor[HTML]{F2F2F2}83  & \cellcolor[HTML]{F2F2F2}24  & \multicolumn{1}{l|}{909}  \\
\multicolumn{1}{|l|}{2}   & 50                         & \cellcolor[HTML]{DDEBF7}32  & \cellcolor[HTML]{DDEBF7}37  & \cellcolor[HTML]{DDEBF7}66  & \cellcolor[HTML]{DDEBF7}81  & \cellcolor[HTML]{DDEBF7}82  & 87                          & 8                          & 87                          & 166                         & 96                          & 75                          & 0                         & 15                          & 83                          & 24                          & \multicolumn{1}{l|}{989}  \\
\multicolumn{1}{|l|}{3}   & 50                         & 16                          & 21                          & 50                          & 65                          & 66                          & 87                          & 8                          & \cellcolor[HTML]{DDEBF7}167 & 166                         & 96                          & 75                          & 0                         & 15                          & 83                          & 24                          & \multicolumn{1}{l|}{989}  \\
\multicolumn{1}{|l|}{4}   & 50                         & 16                          & 21                          & 50                          & 65                          & 66                          & 87                          & 8                          & 87                          & \cellcolor[HTML]{DDEBF7}206 & \cellcolor[HTML]{DDEBF7}116 & \cellcolor[HTML]{DDEBF7}95  & 0                         & 15                          & 83                          & 24                          & \multicolumn{1}{l|}{989}  \\
\multicolumn{1}{|l|}{5}   & 50                         & 16                          & 21                          & 50                          & 65                          & 66                          & 87                          & 8                          & 87                          & 166                         & 96                          & 75                          & 0                         & \cellcolor[HTML]{DDEBF7}95  & 83                          & 24                          & \multicolumn{1}{l|}{989}  \\
\multicolumn{1}{|l|}{6}   & \cellcolor[HTML]{DDEBF7}-  & \cellcolor[HTML]{DDEBF7}-   & \cellcolor[HTML]{DDEBF7}-   & \cellcolor[HTML]{DDEBF7}-   & \cellcolor[HTML]{DDEBF7}-   & \cellcolor[HTML]{DDEBF7}-   & \cellcolor[HTML]{DDEBF7}-   & \cellcolor[HTML]{DDEBF7}-  & \cellcolor[HTML]{DDEBF7}-   & \cellcolor[HTML]{DDEBF7}-   & \cellcolor[HTML]{DDEBF7}-   & \cellcolor[HTML]{DDEBF7}-   & \cellcolor[HTML]{DDEBF7}- & \cellcolor[HTML]{DDEBF7}-   & \cellcolor[HTML]{DDEBF7}-   & \cellcolor[HTML]{DDEBF7}-   & \multicolumn{1}{l|}{989}  \\
\multicolumn{1}{|l|}{7}   & \cellcolor[HTML]{DDEBF7}54 & \cellcolor[HTML]{DDEBF7}17  & \cellcolor[HTML]{DDEBF7}22  & \cellcolor[HTML]{DDEBF7}54  & \cellcolor[HTML]{DDEBF7}70  & \cellcolor[HTML]{DDEBF7}71  & \cellcolor[HTML]{DDEBF7}94  & \cellcolor[HTML]{DDEBF7}8  & \cellcolor[HTML]{DDEBF7}94  & \cellcolor[HTML]{DDEBF7}180 & \cellcolor[HTML]{DDEBF7}104 & \cellcolor[HTML]{DDEBF7}81  & \cellcolor[HTML]{DDEBF7}0 & \cellcolor[HTML]{DDEBF7}16  & \cellcolor[HTML]{DDEBF7}90  & \cellcolor[HTML]{DDEBF7}26  & \multicolumn{1}{l|}{989}  \\ \hline
\end{tabular}
\caption{
The input table for Figure \ref{modified_predictions}. This table records the attempts of urban function adjustment using the computational model under four urban locations. (Columns 00 to 15 annotate the number of 16 urban function types in Table \ref{table:1} as input of each attempt; rows 1 to 7 indicate: numbers of original functions, numbers after adjustment of shopping-related urban functions, adjustment of residential areas, adjustment of workplace-related urban functions, adjustment of public transportation-related urban functions such as bus station, and the case making the number of all types of functions mutually equal to each other, the case maintaining the original proportion of functions)}
\label{table:8}
\end{table*}

\subsection{Combined with Genetic Algorithm}
Further, we embed the trained model into a genetic algorithm. The combination of genetic algorithms and predictive models (Figure \ref{GA}) help discover the optimal or near-optimal solution for our living environment under a specific travel distribution objective. Firstly, a group of urban functional ratios that satisfy the initial constraints is created randomly. Individual objects are entered into a hybrid ANNs model and output predictions, and further evaluations are performed to derive a metric related to travel demand. Referring to this metric as the fitness, the best group of individual objects representing the built environment is selected and subsequently reassessed. The cycle of evaluation, selection, and derivation operations repeats until an optimal or near-optimal solution arrives.\par
Urban decision-makers can set the desired travel demand for a certain urban area or period to solve problems such as the temporal distribution of transportation resources, which might require optimizing the ratios of urban functions within a range for a specific location. In a practical sense, that involves mixing residential and office areas to construct self-contained communities at the proper scale to solve the problem of traffic peaks in job to home.\par
With some locations of interest, the functional ratios in the built environment are used as individual objects for optimization by the genetic algorithm under certain constraints and objectives set by the “designer.” The predictive model or its variants are used as the fitness function to optimize the functional ratios.\par
\begin{figure}[!htb]
    \includegraphics[width=0.9\textwidth]{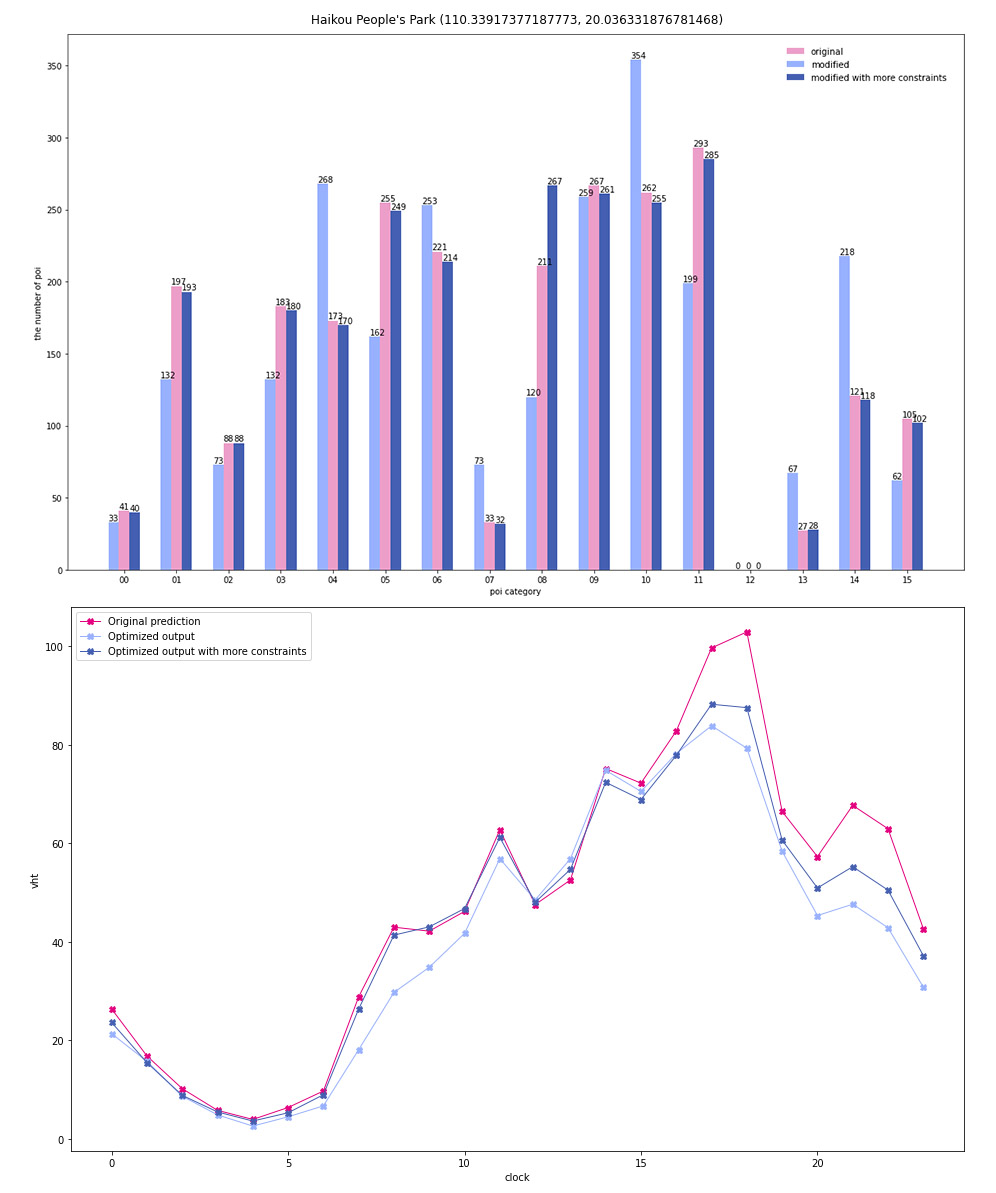}
    \caption{The optimized result in Haikou People's Park}
    \label{opt_layout}
\end{figure}

Using Haikou People's Park as an example, the number of total urban function interest points and type-12 (transportation hub) remain unchanged. The remaining interest points vary up or down by 50 based on the current number, and all are greater than or equal to 0. The objective function is to minimize the variance of vehicle travel demand at different periods to optimize the invocation of transportation resources and release the traffic pressure during peak periods. After the iterative process of the genetic algorithm, the designer can obtain the optimized results as shown in the figure \ref{opt_layout} (light blue line). Indeed, the peak of travel demand can be relieved with the new urban function ratio.\par

In addition, although invoking the trained ANNs model for each prediction takes 525 ms ± 6.41 ms, the genetic algorithm performs many operations, and the process can take up to several hours. The designer can place additional constraints to better fit the actual situation and significantly reduce the optimization time of the genetic algorithm. For example, the designer can set four further base parameters weighted to determine the amount of variation in the four POI categories of eating, housing, work, and public transportation. They can subsequently optimize these four parameters in the genetic algorithm instead of optimizing all 16 POI quantity parameters to conserve time. The figure \ref{opt_layout} also contains the results of this more constrained optimization (dark blue line), which also maps out a more reasonable travel demand time variation with a lower peak than the original one (magenta line). The proposed POI ratios (dark blue columns) derived from this approach are less modified for the built environment. In the case demonstrated here, all functional categories remain largely unchanged except for residential density, which should be increased.\par

\section{Conclusion and Further Study}
Past research revealed an intrinsic link between land use and travel demand, which helped designers consider urban functional planning from a new perspective. However, there lacked a proper methodology allowing the ability designers to obtain detailed travel demand predictions and clear recommendations for urban planning adjustments. Also, the methodology of past studies resulted in a lack of a sufficient basis to migrate findings between different regions.\par
Meanwhile, the emergence of machine learning and big data created a new paradigm of urban research, and several studies utilized machine learning models to enhance traditional workflow and take a more detailed analysis and prediction of traditional urban problems. Our research develops a new computational model to map the relationship between built environments and travel demand, thus creating a network of feedback in the urban planning and design process. The study shows how computational models can be adapted and trained to accurately predict the relationship between an array representing the built environment and vehicle travel demand. The accuracy and transferability of the computational models are further proven to be feasible. With the trained model, we can accurately predict the vehicle travel demand that may occur at a location at various times.\par
With this prediction model, designers can establish explicit urban function compositions as the input and make predictions to obtain a detailed vehicle travel demand profile including total amount and its time distribution. In addition, we show separately in our application exploration how the designer can adjust the inputs to obtain feedback and further leverage the genetic algorithm to obtain explicit suggestions for city function adjustment with the expected travel demand target. In addition, other factors such as socio-economic and policy factors can also affect traffic behavior and travel demand \citep{forsyth2008design, mcnally1997assessment}. In future studies, we aim to quantify these behaviors.\par

\section{Appendix}

\begin{table}[!htb]
\begin{tabular}{|c l l|}
    \hline
    Index & Building Function         &  \\
    \hline
    00    & automobile and motorcycle related  &\\
    \hline
    01    & food and beverages related   & \\
    \hline
    02    & shopping related place      &    \\
    \hline
    03    & daily life service place        &    \\
    \hline
    04    & sports and recreation place     &    \\
    \hline
    05    & medical and health care service place  &     \\
    \hline
    06    & accommodation service related &    \\
    \hline
    07    & tourist attraction related    &   \\
    \hline
    08    & residential area         &      \\
    \hline
    09    & enterprise     &   \\
    \hline
    10    & governmental and social groups related   &\\
    \hline
    11    & science and education cultural place  &\\
    \hline
    12    & traffic hinge & \\
    \hline
    13    & transit network &\\
    \hline
    14    & finance and insurance service institution      &                   \\
    \hline
    15    & public facility    &  \\
    \hline
\end{tabular}
\caption{Urban Function Category According to AutoNavi}
\label{table:1}
\end{table}

\section*{}
\bibliographystyle{apalike}
\bibliography{refs}

\end{document}